\documentclass[conference]{IEEEtran}
\IEEEoverridecommandlockouts
\usepackage{cite}
\usepackage{amsmath,amssymb,amsfonts}
\usepackage{algorithmic}
\usepackage{graphicx}
\usepackage{textcomp}
\usepackage{xcolor}
\def\BibTeX{{\rm B\kern-.05em{\sc i\kern-.025em b}\kern-.08em
    T\kern-.1667em\lower.7ex\hbox{E}\kern-.125emX}}

\usepackage{amsthm}
\usepackage{amsmath,amssymb}
\usepackage{bm}
\usepackage{bbm}

\usepackage{algorithm}
\usepackage{algorithmic}
\usepackage{graphicx}

\usepackage{mathtools}
\mathtoolsset{showonlyrefs,showmanualtags}
\usepackage{bbm}

\usepackage{multirow}
\usepackage{booktabs}

\usepackage{caption}
\usepackage{subcaption}

\usepackage{url}
\usepackage{hyperref}

\usepackage{lipsum}

\AtBeginDocument{
  \abovedisplayskip     =0.2\abovedisplayskip
  \abovedisplayshortskip=0.2\abovedisplayshortskip
  \belowdisplayskip     =0.2\belowdisplayskip
  \belowdisplayshortskip=0.2\belowdisplayshortskip}

\renewcommand{\arraystretch}{0.85}

\begin{document}

%\title{Graph Novelty Generation with GMM-based Diffusion Models}
\title{Graph Community Augmentation\\with GMM-based Modeling in Latent Space}
%\title{Graph Community Augmentation\\from a Viewpoint of MDL}

\author{\IEEEauthorblockN{Shintaro Fukushima}
\IEEEauthorblockA{\textit{Social PF Development Division} \\
\textit{Toyota Motor Corporation}\\
Tokyo, Japan \\
%sfukushim@gmail.com}
shi-fukushima@toyota-tokyo.tech}
\and
\IEEEauthorblockN{Kenji Yamanishi}
\IEEEauthorblockA{\textit{Graduate School of Information Science and Technology} \\
\textit{The University of Tokyo}\\
Tokyo, Japan \\
yamanishi@g.ecc.u-tokyo.ac.jp}
}
%\author{\IEEEauthorblockN{Anonymous}
%    \IEEEauthorblockA{\textit{Anonymous}}
%}

\maketitle

\theoremstyle{definition}
\newtheorem{definition}{Definition}[section]
\newtheorem{theorem}{Theorem}[section]
\newtheorem{assumption}[theorem]{Assumption}
\newtheorem{proposition}[theorem]{Proposition}
\newtheorem{lemma}[theorem]{Lemma}
\newtheorem{corollary}[theorem]{Corollary}

\newtheorem{example}{Theorem}[section]

\let\Bbbk\relax

%----------------------------------------
% algorithm
%----------------------------------------

\renewcommand{\algorithmicrequire}{\textbf{Input:}}
\renewcommand{\algorithmicensure}{\textbf{Output:}}

%----------------------------------------
% equation
%----------------------------------------

% mydef
\newcommand*{\MyDef}{\mathrm{def}}
\newcommand*{\eqdefU}{\ensuremath{\mathop{\overset{\MyDef}{=}}}}% Unscaled version
\newcommand*{\eqdef}{\mathop{\overset{\MyDef}{\resizebox{\widthof{\eqdefU}}{\heightof{=}}{=}}}}
\newcommand\mydef{\mathrel{\overset{\makebox[0pt]{\mbox{\normalfont\tiny\sffamily def}}}{=}}}

% vector
\renewcommand{\vec}[1]{\boldsymbol{#1}}
% transpose
\def\transpose#1{#1^{\top}}
% argmax
\newcommand{\argmax}{\operatornamewithlimits{argmax}}
% argmin
\newcommand{\argmin}{\operatornamewithlimits{argmin}}
% for integral
\def\d#1{\mathrm{d}#1}

\makeatletter
\def\env@cases{%
  \let\@ifnextchar\new@ifnextchar
  \left\lbrace
  \def\arraystretch{1.0}%
  \array{l@{\quad}l@{}}% Formerly @{}l@{\quad}l@{}
}

\thickmuskip=0.5\thickmuskip
\medmuskip=0.5\medmuskip
\thinmuskip=0.5\thinmuskip

\begin{abstract}
This study addresses the issue of graph generation with generative models. 
In particular, we are concerned with \textit{graph community augmentation} problem, 
which refers to the problem of generating \textit{unseen} or \textit{unfamiliar} graphs with a new community 
out of the probability distribution estimated with a given graph dataset. 
The graph community augmentation means that the generated graphs have a new community. 
There is a chance of discovering an unseen but important structure of graphs with a new community, for example, 
in a social network such as a purchaser network. 
Graph community augmentation may also be helpful for generalization of data mining models in a case where it is difficult to collect real graph data enough. 
In fact, there are many ways to generate a new community in an existing graph. 
It is desirable to discover a new graph with a new community beyond the given graph 
while we keep the structure of the original graphs to some extent 
for the generated graphs to be realistic. 
To this end, 
we propose an algorithm  
called the graph community augmentation~(GCA). 
The key ideas of GCA are (i) to fit Gaussian mixture model~(GMM) to data points in the latent space into which the nodes in the original graph are embedded, 
and 
(ii) to add data points in the new cluster in the latent space for generating a new community 
based on the minimum description length~(MDL) principle. 
We empirically demonstrate the effectiveness of GCA for generating graphs with a new community structure on synthetic and real datasets. 
\end{abstract}


\section{Introduction}
\label{section:introduction}

\subsection{Motivation}
\label{subsection:motivation}
In recent years, 
we have witnessed the great success of generative models  
in the areas of knowledge discovery and data mining 
for various domains, 
such as images, texts, audios, time-series, and graphs~\cite{Bond-Taylor2022TPAMI}. 
The generative models aim at learning probabilistic distribution underlying a given dataset. 
%and have a wide range of applications including 
%image and text generation, 
%molecular science, drug discovery, and material science.  
Graph generation with generative models is one of the ever-evolving fields in the research line of generative models~(e.g., \cite{Bonifati2020CSUR,Faez2021IEEEAccess,Guo2022TPAMI,Zhu2022LoG}). 
%Currently, 
%the diffusion models~\cite{Sohl-Dickstein2015,Ho2020NeurIPS} are the state-of-the-art generative models, 
%which have shown its powerful expressiveness 
%with the stability in training 
%as well as 
%the amazing ability to generate data~\cite{Yang2023ACS}. 

\begin{figure}[tb]
\centering
\includegraphics[width=\linewidth]{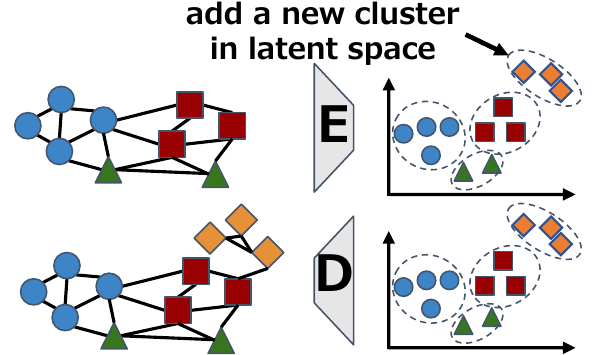}
%\caption{Illustration of the graph novelty generation problem. 
\caption{Illustration of the graph community augmentation problem. 
``E'' denotes an encoder that maps nodes in a graph into data points in a latent space, 
whereas 
``D'' denotes a decoder that maps data points in a latent space into a new graph. 
%The nodes in the original graph are embedded into a latent space with an encoder, and new clusters are added in the space. 
%These points are diffused into random noises in the diffusion process. Then, these noises are denoised in the generative~(denoising) process and decoded into a graph with a decoder.
}
\label{fig:illustration_of_the_graph_novelty_generation_problem}
\end{figure}

%In this study, we address the \textit{graph novelty generation} problem with diffusion models. 
In this study, we address the \textit{graph community augmentation} problem. 
This problem refers to 
%generating \textit{unseen} or \textit{unfamiliar} graphs  
generating \textit{unseen} or \textit{unfamiliar} graphs with a community 
out of the probability distribution estimated with a given dataset. 
Note that the graph community augmentation is quite different from the conventional \text{graph data augmentation}~(e.g., \cite{Ding2022SIGKDDExplorations,Zhao2023IDEB}) in the sense that the former means generating graphs with a new community structure, 
while the latter means manipulating one or more types of graph structures, 
such as node attributes, 
node labels, and edge connection. 
%The novelty means that the generated graphs have novel structure or property in relation to their inherent nature. 
%In particular, we consider the problem with diffusion models 
%for leveraging their powerful expressiveness. 
%Fig.~\ref{fig:illustration_of_the_graph_novelty_generation_problem} illustrates the graph novelty generation problem; 
Fig.~\ref{fig:illustration_of_the_graph_novelty_generation_problem} illustrates the graph community augmentation problem; 
%we first learn probabilistic structure of a given graph dataset with diffusion models in a latent space, 
we first learn probabilistic structure of a given graph dataset in a latent space, 
%and then add new components in the latent space. 
and then add a new component in the latent space. 
%These points are diffused into random noises in the diffusion process. 
%Then, these noises are denoised in the reverse~(denoising) process and decoded into a graph with a decoder. 
%One might wonder why the diffusion model~(diffusion process and reverse process) is necessary. 
%In other words, it might be reasonable to build only the encoder and decoder, to generate data points in the latent space and decode them to generate a new graph. 
%However, with such an approach, noises~(outliers) are often contaminated. 
%The diffusion model has its strong expressiveness as well as denoising ability. 
%We can generate novel and reliable graphs by leveraging the diffusion model. 

%Thus far, there have been several lines of research related to the novelty generation problem; 
%deep graph generation~\cite{Guo2022TPAMI,Zhu2022LoG,Faez2021IEEEAccess,Bonifati2020CSUR}, 
%\textit{extrapolation problem} 
%of the generative models~~\cite{Feng2021ICML,Chan2021NeurIPS,Besserve2021AAAI,Li2021WACV,Han2019KDD}, 
%graph data augmentation~\cite{Ding2022SIGKDDExplorations,Zhao2023IDEB}, 
%out-of-distribution~(OOD) distribtuion~\cite{Li2022arXiv}, 
%and counterfactual synthesis. 
%However, 
%their main purpose is to learn a probabilistic distribution and then to generate graphs from the distribution, 
%or even when their purpose is to generate data outside of the distribution, 
%their tasks and data are different from our graph novelty generation problem. 

However, 
%why is the graph novelty generation problem important? 
why is the graph community augmentation problem important? 
It is because there is a chance of discovering an unseen but important structure of graphs 
with a new community 
beyond a dataset for learning its probabilistic structure~(e.g., \cite{Ohsawa1998ADL,Ohsawa2002NGC}). 
For example, 
in a social network, 
such as a purchaser network, 
a new but closely related cluster might indicate a sign of chance or a new emergence of knowledge~(a new class or group of purchasers) in the near future. 
Graph community augmentation may also be helpful for generalization of data mining models in a case where it is difficult to collect real graph data enough. 
%Another example is in the domain of molecular discovery. 
%A new novel part added by the graph novel generation might show effectiveness in their material properties. 
To this end, 
%it is desirable to discover new graphs out of the distribution 
it is desirable to discover new graphs with a new community out of the distribution 
while we keep the structure of the original graphs to some extent for the generated graphs to be realistic. 
Such an ability to generate novel graphs is not realizable with current graph generative models. 
%Therefore, our research question is summarized as follows: \textit{how can we efficiently generate novel graphs while keeping basic structure of a given graph dataset?} 
%Therefore, 
Our research question is summarized as follows: \textit{how can we efficiently generate novel graphs with a new community structure while keeping basic structure of a given graph dataset?} 
We answer this question 
%with diffusion models 
from an information-theoretic perspective, 
in particular, 
based on the minimum description length~(MDL) principle. 

\subsection{Significance and Novelty}
\label{subsection:significance_and_novelty}

The significance and novelty of this study are summarized as follows:
\begin{enumerate}
%\item \textit{A new formulation for graph novelty generation problem.}
\item \textit{A new formulation for graph community augmentation problem.}
The key of our approach is to embed the original graph into the latent space and to represent it with a GMM (Gaussian Mixture Model). Here each cluster in the GMM may correspond to a substructure of the original graph such as a community. 
%We realize graph novelty generation by decoding of a new cluster in the GMM. 
We realize graph community augmentation by decoding of a new cluster in the GMM. 
%The important issue is here how we should put a new cluster in the GMM. 
The important issue here is how we should put a new cluster in the GMM. 
We mathematically impose the novelty and reliability conditions on it. 
%We employ the diffusion model to generate graph novelty that is associated with a significantly different from others but does not change the overall distributions so much. 
We employ MDL to generate graph novelty that is associated with a significantly different from others but does not change the overall distributions so much. 
%Thus this research gives the first formulation for novel graph generation.
Thus, this research gives the first formulation for graph community augmentation. 
It is also  the first work on  knowledge extrapolation in terms of graph communities.

%\item \textit{A novel algorithm for addressing the novelty graph generation problem.}
\item \textit{A novel algorithm for addressing the graph community augmentation.}
We give a concrete algorithm for implementing the ideas in 1). 
%It consists of training phase and the novelty generating one. 
It consists of training phase and the community augmentation one. 
In the former, we employ the autoencoder to embed the original graph into the latent space and conduct model selection of the number of clusters in GMM on the basis of the MDL principle. 
In the latter, 
we allocate a new cluster~(community) by considering the novelty and reliable conditions. 
%The novelty condition assigns a constraint that a new cluster is apart from the existing clusters, 
%while the reliable condition assigns a constraint that the new probabilistic probability distribution along with the new cluster is not so different from the original distribution. 
Specifically, 
we define the novelty condition with KL`(Kullback-Leibler) divergence between the newly-added and existing clusters, 
while we define the reliability condition with KL divergence between the newly-generated and original probability distributions. 
%we employ the GMM-based diffusion model. Specifically, in the denoising process, the EM algorithm and diffusion process are conducted jointly with MDL model selection under the novelty and reliable conditions for a new component. 
%This is the first application of GMM-based diffusion model to novelty graph generation.
%This is the first application to graph community augmentation.

\item \textit{Empirical demonstration of the effectiveness of the proposed algorithm on both synthetic and real graph datasets.} 
The effectiveness of our methodology is quantitatively evaluated in comparison with existing graph generative models, in terms of the degree of maintenance of structure of original
graphs, and the degree of extrapolation search. 
We empirically demonstrate that our method is able to generate novel graphs while keeping the essential nature of the original graphs in a reasonable computation time. 
\end{enumerate}

\section{Related Work}
\label{section:related_work}

In this section, 
we briefly review the related work. 
Our study is related to graph generation, embedding with GMM, extrapolation problem for generative models, and other somewhat closely related topics, 
such as 
graph data augmentation, 
OOD~(out-of-distribution) 
generalization, 
and counterfactual synthesis. 

Graph generation aims at learning generative models. 
In addition to retrospective models such as ERGMS~\cite{Harris2014book}, 
approaches with deep neural network 
have been extensively explored, recently. 
%in the areas of knowledge discovery and data mining as well as machine learning
See comprehensive surveys for details~\cite{Guo2022TPAMI,Zhu2022LoG,Faez2021IEEEAccess,Bonifati2020CSUR}. 
%%such as images, texts, audios, time-series data, 
%and graphs. 
%Thus far, the proposed generative models 
%include autoregressive models~\cite{Oord2016ICML}, 
%variational autoencoder~(VAE)~\cite{Kingma2014ICLR,Rezende2014ICML}, 
%generative adversarial network~(GAN)~\cite{Goodfellow2014NIPS}, 
%energy-based models~\cite{Lecun2006PSD,Ngiam2011ICML}, 
%and normalizing flow~\cite{Papamakarios2021JMLR,Rezende2015ICML,Kobyzev2021TPAMI}. 
%The diffusion model is one of the brand-new generative models~\cite{Sohl-Dickstein2015,Ho2020NeurIPS}, 
%due to their stability in training, 
%realization of conditional data generation, 
%and easiness of incorporation of symmetry and invariance in data generation. 
%Several studies have applied the diffusion models  
%to graph generation~\cite{Niu2020AISTATS,Austin2021NeurIPS,Hoogeboom2021NeurIPS,Huang2022ICDM,Jo2022ICML,Konstantin2022LoG,Vignac2023ICLR,Kong2023ICML,Chen2023ICML,Jo2023arXiv}. 
%%See also comprehensive surveys for overview~\cite{Liu2023IJCAI,Fan2023arXiv,Zhang2023arXiv}. 
%See comprehensive surveys~\cite{Liu2023IJCAI,Fan2023arXiv,Zhang2023arXiv}. 
However, 
%despite the efforts in the previous studies, 
%most of the studies have focused only on learning the underlying distribution from a training dataset. 
most of the studies focus only on learning the underlying distribution from a training dataset. 
%Therefore, these studies have not addressed the issue of generating data outside of the distribution estimated. 
Therefore, these studies do not address the issue of generating data outside of the estimated distribution. 
%the \textit{novelty generation} of graphs with generative models. 

Several studies have addressed the issue of embedding a graph into a latent space with GMM. 
ComE~\cite{Cavallari2017CIKM} learns community embedding in combination with community detection and node embedding. 
DGG~\cite{Yang2019ICCV} clusters nodes with graph embedding. 
However, these studies only detect community or cluster nodes in a graph in a latent space. 
Therefore, the algorithms proposed there do not generate a new community. 

The extrapolation problem for generative models 
has been addressed in several studies~\cite{Feng2021ICML,Chan2021NeurIPS,Besserve2021AAAI,Li2021WACV,Han2019KDD}. 
Although some of these studies have sound theoretical foundations, 
they focus on tasks and data different from our graph community augmentation; 
sequence generation for natural language processing and proteins~\cite{Chan2021NeurIPS}, 
search extrapolation and recommendation~\cite{Han2019KDD}, 
and 
expanding image field of view~\cite{Li2021WACV}. 
Besides, 
the generative models employed there are GAN~\cite{Besserve2021AAAI,Feng2021ICML,Li2021WACV}, 
autoregressive model~\cite{Chan2021NeurIPS}, 
and 
Sequence-to-Sequence~\cite{Han2019KDD}. 
%Therefore, no study has applied the diffusion models to the  extrapolation problem. 
%No study has addressed the issue of generating a new community. 

There are also other lines of research related to ours, 
such as 
graph data augmentation, 
%data corruption, 
OOD %(out of distribution)
generalization on graphs, 
and counterfactual synthesis. 
First, 
the graph data augmentation~\cite{Ding2022SIGKDDExplorations,Zhao2023IDEB} has been extensively studied to enhance the ability of generalization of graph learning, 
with (semi-)~supervised learning~\cite{Azabou2023ICML,Liu2022ICML,Li2023ICDM} 
as well as 
self-supervised learning~\cite{Lin2023ICLR,Kim2022NeurIPS}. 
However, 
the generated data %with graph data augmentation 
are limited in that they are generated from the underlying distribution and thus have little novelty. 
%Second, 
%data corruption has attracted attention, recently~(e.g., \cite{Daras2023NeurIPS,Elata2024TMLR}). 
%However, the primary concern of the data corruption is 
%to effectively train models by destroying the existing data structure with added noises. 
%Third, 
Second, 
OOD generalization on graphs~\cite{Li2022arXiv} aims at generalization of classification models when distributions of training and test datasets are different, and thus also is a somewhat closely related topic. 
%For example, 
%Lee et al.~\cite{Lee2023ICML} proposed a score-based diffusion scheme that incorporates OOD control in the generative stochastic differential equation~(SDE) with simple control of a hyper-parameter. 
However, OOD generalization aims at adapting to the test dataset, and the primary purpose is not extrapolation. 
Finally, 
counterfactual synthesis is one of the promising approaches for knowledge extrapolation~\cite{Kocaoglu2018ICLR,Sauer2020ICLR,Nemirovsky2020arXiv,Yang2021CVPR,Averitt2020JBiomedInform,Thiagarajan2021NeurIPS}. 
Most of these studies directly model a structural causal model~\cite{Pearl2009StatSurveys}.  
However, 
the purpose of counterfactual synthesis is to generate data samples for changing outputs, 
for example, classification or regression results. 

In summary, despite extensive efforts in the previous studies mentioned above, 
to the best of our knowledge, 
%there has been no study that leverages the diffusion models to generate novelty graphs efficiently and effectively. 
there has been no study that generates a new community structure efficiently and effectively.

\section{Problem Setting}
\label{section:problem_setting}

In this section, 
we provide information related to this study 
and our problem setting.  
We assume that a directed graph $G = (X, \, A)$ is given, 
where 
%$V$ is a set of nodes 
%and the number of nodes is $N$; $|V|=N$, 
$X \in \mathbb{R}^{N \times F}$ is a feature matrix of the nodes, 
$A \in \mathbb{R}^{N \times N}$ is a weighted adjacency matrix between the nodes; 
$N \in \mathbb{N}$ is the number of nodes 
and 
$F \in \mathbb{N}$ is the number of features. 
%It is one of the future work to consider a case where a set of node labels $y \in \{1, \, \dots, \,  K\}^{N}$ is given to a graph, where $K \in \mathbb{N}$ is the number of classes. 
%We also consider a case where a set of node labels $Y \in \{1, \dots, C\}$ is given, where $C \in \mathbb{N}$ is the number of classes.  
%That is, 
%$G = (X, Y, A)$. 
%However, this case is only an option. 

%\subsection{Problem Formulation}
%\label{subsection:problem_formulation}

Our problem setting is stated as follows. 
Note that this section gives a brief sketch of our problem setting, without stating exact or concrete expressions for all the notations. 
Their exact definitions will be given in the following sections. 

\subsection{Overall Framework}
\label{subsection:overall_framework}

First, we map a set of nodes in $G$ into a set of data points in a latent space with a trained encoder:
\begin{align}
\bm{v} = \mathrm{Encoder}(G), 
\label{eq:encode_a_graph_into_lantent_representation}
\end{align}
where $\bm{v} = \{ v_{i} \}_{i=1}^{N} \in \mathbb{R}^{N \times D}$ is a set of data points in the latent space, 
$v_{i} \in \mathbb{R}^{D}$ is a data point in a $D$-dimensional latent space, 
which corresponds to the $i$th node in $G$~($D \in \mathbb{N}$). 
$\mathrm{Encoder}$ is a trained encoder, 
the concrete expression of which is given in Sec.~\ref{subsubsection:graph_autoencoder}. 
For a training dataset, 
we learn a probabilistic density of data points 
%$\bm{v} \in \mathbb{R}^{N \times D}$. 
$\bm{v}$. 
There are many ways for density estimation. 
In this study, 
we use GMM~\cite{Bishop2006PRML} 
because it approximates all the density functions 
and it is easy to add new clusters or remove the existing ones with GMM, 
which is suitable for the community generation problem. 
The probabilistic density of $v_{i}$ is expressed as 
\begin{align}
&p_{\mathrm{orig}}(v_{i}; \{ \mu_{k} \}_{k=1}^{K}, \{ \Sigma_{k} \}_{k=1}^{K})
= \sum_{k=1}^{K} w_{k} \mathcal{N}(v_{i}; \mu_{k}, \Sigma_{k}), 
\, 
\sum_{k=1}^{K} w_{k} = 1, \label{eq:original_probability_density}
\end{align}
where $\mu_{k} \in \mathbb{R}^{D}$ 
and $\Sigma_{k} \in \mathbb{R}^{D \times D}$ 
denote 
the mean and variance-covariance matrix 
of the $k$-th component, 
respectively. 
Then, we consider adding new  components for generating novel graphs. 
If we add 
%k_{\mathrm{new}}$ clusters~($k_{\mathrm{new}} \in \mathbb{N}$), 
a cluster, 
the probabilistic density of $v_{i}$ is expressed as follows:
\begin{align}
&p_{\mathrm{new}}(v_{i}; 
  %\{ \mu_{k} \}_{k=1}^{K + k_{\mathrm{new}}},
  %\{ \Sigma_{k} \}_{k=1}^{K + k_{\mathrm{new}}})
  \{ \mu_{k} \}_{k=1}^{K+1},
  \{ \Sigma_{k} \}_{k=1}^{K+1})
= 
%\sum_{k=1}^{K+k_{\mathrm{new}}} 
\sum_{k=1}^{K+1} 
  w_{k}' \mathcal{N}(v_{i}; \mu_{k}, \Sigma_{k}), \\
%\sum_{k=1}^{K+k_{\mathrm{new}}}  w_{k}' &= 1,
&\sum_{k=1}^{K+1}  w_{k}' = 1,
\label{eq:probability_density_with_newly_generated_clusters}
\end{align}
%
%\begin{align}
%&p_{\mathrm{new}}(v; \{ \mu_{k} \}_{k=1}^{K-1} \oplus \{ \mu_{K} \}, 
%\{ \Sigma_{k} \}_{k=1}^{K-1} \oplus \{ \Sigma_{K} \})  \\
%&= \sum_{k=1}^{K-1} w_{k}' \mathcal{N}(v; \mu_{k}, \Sigma_{k})
%+ w_{K}' \mathcal{N}(v; \mu_{K}, \Sigma_{K}), \, 
%\sum_{k'=1}^{K} w_{k'}' = 1. 
%\end{align}
where $\{ \mu_{k} \}_{k=1}^{K}$ and $\{ \Sigma_{k} \}_{k=1}^{K}$ are the sets of the existing means and variance-covariance matrices in Eq.~\eqref{eq:original_probability_density}, respectively. 
By contrast, 
%$\{ \mu_{k} \}_{k=K+1}^{K + k_{\mathrm{new}}}$ 
$\mu_{K+1}$ 
and 
%$\{ \Sigma_{k} \}_{k=K+1}^{K+k_{\mathrm{new}}}$ 
$\Sigma_{K+1}$ 
are a newly added 
%set of 
mean and variance-covariance matrix, respectively. 
%Then, 
%we diffuse data points $\bm{h}^{0}$ according to Eq.~\eqref{eq:diffusion_process_explicit} 
%in the diffusion process,  
%and 
%%denoise $h^{t}$ and thus obtain $h^{t-1}$ by learning $p_{\theta}(h^{t-1} | h^{t})$ 
%denoise $h^{t}$ to obtain $h^{t-1}$ by learning $p_{\theta}(h^{t-1} | h^{t})$ 
%in Eq.~\eqref{eq:parameterized_conditional_probability_density_in_reverse_process} 
%in the reverse process. 
%By repeating this procedure, 
%we obtain $\bar{\bm{h}}^{0} = \{ \bar{h}_{i}^{0} \}_{i=1}^{N} \in \mathbb{R}^{N \times D}$. 
%Note that this procedure is conducted for generating categorical data with the diffusion models and GMM~\cite{Regol2023AAAI}. 
%We then decode $\bar{\bm{h}}^{0}$ into a graph: 
We then sample %a data points 
data points 
$\bm{v}' \in \mathbb{R}^{M \times D}$ from $\mathcal{N}(v; \mu_{K+1}, \Sigma_{K+1})$ 
with probability $w_{K+1}$~($M \in \mathbb{N}$). 
We concatenate $\bm{v}$ and $\bm{v}'$ 
to a matrix $\bm{v} \| \bm{v}' \in \mathbb{R}^{(N+M) \times D}$ 
and decode 
%$\bm{v} \oplus \bm{v}' \in \mathbb{R}^{(N+M) \times D}$ into a graph: 
%$\bm{v} \| \bm{v}' \in \mathbb{R}^{(N+M) \times D}$ into a graph: 
it into a graph: 
\begin{align}
G' &= \mathrm{Decoder}(
    %\bm{v} \oplus \bm{v}'
    \bm{v} \| \bm{v}'
), 
\label{eq:decode_latent_representation_into_a_graph}
\end{align}
where $\mathrm{Decoder}$ is a trained decoder, 
the concrete expression of which is given in Sec.~\ref{subsubsection:graph_decoder}. 
%We train the $\mathrm{Encoder}$ and $\mathrm{Decoder}$ 
%so that the loss between $G$ and $G'$ is minimized. 
 
%As we mentioned it in Sec.~\ref{section:introduction}, 
%note that the diffusion model~(diffusion process and reverse process) plays an essential role in generating novel and reliable graphs. 

\subsection{Reliable and Novel Data Generation}
\label{subsection:reliable_and_novel_data_generation}

The desirable property of the generated graph $G'$ is novel, but not so different from the existing graphs. 
%To the end, our algorithm have to meet the following conditions:
To this end, our algorithm has to satisfy the following conditions:
\begin{description}
\item[(i)] \textbf{novelty condition}: the algorithm extrapolates the domain of graphs to generate the novelty. 
\item[(ii)] \textbf{reliability condition}: the algorithm maintains structure of the original graph to some extent for the generated graph to be realistic. 
\end{description}
For (i), the new cluster is placed as far away as possible from the existing ones. 
By contrast, 
for (ii), the generated graph keeps the structure of the original graph. 

%In the following, 
%we formulate the two properties quantitatively. 
%We abbreviate the probabilistic densities in Eq.~\eqref{eq:original_probability_density} and 
%\eqref{eq:probability_density_with_newly_generated_clusters} as $p_{\mathrm{orig}}$ and $p_{\mathrm{new}}$, respectively, 
%as follows: 
%\begin{align}
%p_{\mathrm{orig}} &\mydef 
%p(h^{0}; 
%  \{ \mu_{k} \}_{k=1}^{K}, 
%  \{ \Sigma_{k} \}_{k=1}^{K}
%)
%= \sum_{k=1}^{K}
%    w_{k} \mathcal{N}(h^{0}; \mu_{k}, \Sigma_{k}),
%\,  
%\sum_{k=1}^{K} w_{k} = 1, \\
%p_{\mathrm{new}} &\mydef 
%p(h^{0}; 
%  \{ \mu_{k} \}_{k=1}^{K+k_{\mathrm{new}}},
%  \{ \Sigma_{k}
%  \}_{k=1}^{K+k_{\mathrm{new}}})
%= \sum_{k=1}^{K+k_{\mathrm{new}}} 
%w_{k}' \mathcal{N}(h^{0}; \mu_{k}, \Sigma_{k}), 
%\, 
%\sum_{k=1}^{K+k_{\mathrm{new}}} 
%w_{k}' = 1.
%\end{align}

Then, we consider how (i) and (ii) are formulated mathematically. 
The former condition (i) is to locate each pair of components in the original and newly-generated probability densities as far as possible 
for $k=1, \dots, K$, 
as follows: 
%It is formulated as follows: 
\begin{align}
&d(\mathcal{N}(v; \mu_{k}, \, \Sigma_{k}), \,\,
  \mathcal{N}(v; \mu_{K+1}, \,  \Sigma_{K+1})) 
\geq \delta_{0}, 
%\,\,\,   
%&(k'=K+1, \dots, K+k_{\mathrm{new}}, \,  
%  k=1, \dots, K), 
%(k=1, \dots, K), 
\label{eq:condition_between_existing_and_newly_added_clusters}
\end{align}
where 
$d$ is a (quasi-)distance between probabilistic densities, 
$\delta_{0} \in \mathbb{R}^{+}$ is a specified hyper-parameter 
for threshold of the distance between the existing clusters and the newly added one. 
We call Eq.~\eqref{eq:condition_between_existing_and_newly_added_clusters} 
the \textit{novelty condition}. 
Eq.~\eqref{eq:condition_between_existing_and_newly_added_clusters} indicates that 
each pair of the existing and newly-added clusters 
is more than or equal to $\delta_{0}$. 
Therefore, 
the novelty condition measures how novel the newly-added cluster is relative to existing ones. 

By contrast, 
the latter condition (ii) is realized by locating the newly-generated probabilistic density within a distance from the original one. 
It is formulated as follows: 
\begin{align}
d(p_{\mathrm{orig}}, \,  p_{\mathrm{new}}) &\leq \delta_{1}, 
\label{eq:condition_between_existing_and_newly_generated_probability_densities}
\end{align}
where 
$p_{\mathrm{orig}}$ and 
$p_{\mathrm{new}}$ 
denote the original and newly-generated probability densities 
in Eq.~\eqref{eq:original_probability_density} 
and 
Eq.~\eqref{eq:probability_density_with_newly_generated_clusters}, 
respectively, 
and 
$\delta_{1} \in \mathbb{R}^{+}$ is a specified hyper-parameter for threshold of the distance between the original and newly-generated probabilistic densities. 
We call Eq.~\eqref{eq:condition_between_existing_and_newly_generated_probability_densities} the \textit{reliability condition}. 
The reliability condition measures how similar the original and newly-generated GMMs are. 
Therefore, 
Eq.~\eqref{eq:condition_between_existing_and_newly_generated_probability_densities} indicates that the original and newly-generated GMMs are not different in a resolution of $\delta_{1}$. 

We have to answer the following two questions: 
\begin{enumerate}
%\item[Q1.] How should we  learn the probabilistic density in Eq.~\eqref{eq:original_probability_density} with GMM and diffusion model? 
\item[Q1.] How should we  learn the probabilistic density in Eq.~\eqref{eq:original_probability_density} with GMM? 
\item[Q2.] How should we quantitatively formulate the distance conditions in Eq.~\eqref{eq:condition_between_existing_and_newly_added_clusters} and Eq.~\eqref{eq:condition_between_existing_and_newly_generated_probability_densities}, 
and solve these inequalities? 
\end{enumerate}
%Q1 is related to training GMM and diffusion model, 
Q1 is related to training the autoencoder and GMM, 
whereas Q2 is concerned with data generation by modifying the probabilistic structure of GMM in the latent space in the data generation process. 
We describe the solutions to these two questions 
in the next section. 

%\textcolor{red}{Our problem setting is stated as follows: 
%for a give $\delta \in \mathbb{R}^{+}$, 
%solve the following reverse process: 
%\begin{align}
%p(x_{t-1} | x_{t}) 
%=
%\begin{cases}
%\sum_{i=1}^{M} w_{i} \mathcal{N}(x_{T-1} | \mu_{i}(x_{T}), \Sigma_{i}(x_{T})) & (t=T),  \\
%\sum_{i=1}^{M} w'_{i} \mathcal{N}(x_{t-1} | \mu_{i}(x_{T-1} \oplus y_{T-1}), 
%\Sigma_{i}(x_{T-1} \oplus y_{T-1})) & (1 \leq t \leq T-1), 
%\end{cases}
%\end{align}
%where $y_{T-1}$ is a dataset of noises added at $t=T-1$, 
%satisfying the following equation: 
%\begin{align}
%L(x_{T-1} \oplus y_{T-1}) - L(x_{T-1}) < \delta, 
%\end{align}
%where $L$ is a code-length satisfying Kraft's inequality. 
%}

%\subsection{Problem Formulation~2(Levy process)~\cite{Yoon2023NeurIPS}}

\section{Proposed Algorithm}
\label{section:proposed_algorithm}

\begin{figure}[tb]
\centering
\includegraphics[width=\linewidth]{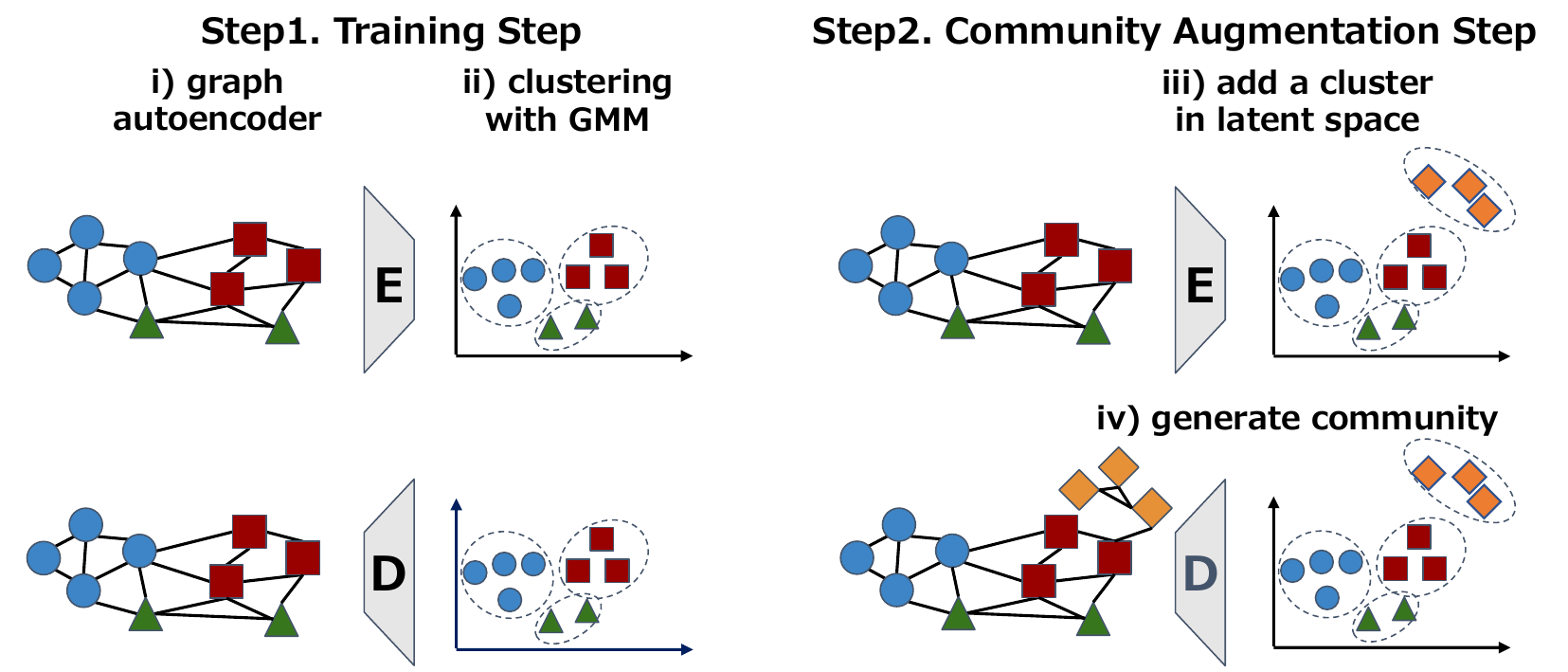}
%\caption{Oveall flow of the proposed algorithm called the novel graph generation with diffusion models~(NGG-DM).}
\caption{Overall flow of the proposed algorithm called the graph community augmentation~(GCA).}
\label{fig:overall_flow_of_proposed_algorithm}
\end{figure}

In this section, 
%we propose an efficient algorithm for graph novelty generation, called the novel graph generation with diffusion models~(NGG-DM). 
we propose an efficient algorithm for graph community augmentation~(GCA). 
Fig.~\ref{fig:overall_flow_of_proposed_algorithm} shows the overall flow of GCA, 
%which is composed of 1) training step and 2) novelty generation step.   
which is composed of 1) training step and 2) community augmentation step.   
In the training step, 
we encode a given graph into  latent representations of nodes with a graph autoencoder, 
then we cluster these representations. 
%and diffuse them %these representations 
%in the diffusion process 
%and reverse them to the original representations in the generative process. 
By contrast, 
%in the novelty generation step, 
in the community augmentation step, 
we add %new clusters 
%a new cluster in the latent space and generate a new graph through the diffusion and generative processes. 
a new cluster in the latent space and generate a new graph with a new community through the generative process. 
%In the following subsections, 
We explain the two steps in detail. 

\subsection{Training Step}
\label{subsection:training_step}

In the training step, 
%there are the following four components: 
there are the following two components: 
%(i) graph autoencoder~(encoder and decoder), 
%(ii) clustering with GMM, 
%(iii) diffusion process, 
%and 
%(iv) %efficient 
%reverse process by leveraging GMM. 
(i) graph autoencoder~(encoder and decoder) 
and 
(ii) clustering with GMM. 
%We describe these steps in detail. 

\subsubsection{Graph Autoencoder}
\label{subsubsection:graph_autoencoder}

We consider how to realize Eq.~\eqref{eq:encode_a_graph_into_lantent_representation} 
and 
Eq.~\eqref{eq:decode_latent_representation_into_a_graph} 
with an autoencoder. 
The autoencoder is a neural network structure for mapping a given datum into a low-dimensional vector in a latent space. 
For a graph, several autoencoder algorithms have been proposed~\cite{Zhang2022TKDE}. 
%Many of existing algorithms are \textit{inductive}, 
%that is, 
%they embed existing nodes and edges in an original graph~(e.g., \cite{Kipf2016arXiv,Wang2016KDD}). 
%However, in our study, 
%it is important to run \textit{transductive} node embedding. 
%In other words, we have to embed unknown nodes and edges in training. 
%To this end, 
%it is a choice to select transductive algorithms such as GraphSAGE~\cite{Hamilton2017NIPS}. 
In this study, we simply employ the variational graph autoencoder~(VGAE)~\cite{Kipf2016arXiv}. 
In that way, 
we obtain a set of embedding vectors $\bm{v} = \{v_{i} \}_{i=1}^{N} \in \mathbb{R}^{N \times D}$ 
for the nodes in the original graph, 
where $D$ is the embedding dimension. 
It is described in Sec.~\ref{subsubsection:graph_decoder} to decode data points in the latent space into a graph. 

\subsubsection{Clustering with GMM}
\label{subsubsection:clustering_with_GMM}

Next, we aggregate the node embedding vectors with GMM. 
For a given number of clusters $k$, 
we assume that each data point belongs to one of the clusters $\{1, \dots, k \}$; 
$z_{i} \in \{ 1, \dots, k \}$ for $i=1, \dots, N$. 
We denote the latent variable for aggregating the membership as $\bm{z} = \{ z_{i} \}_{i=1}^{N}$. 
We assume that the data points in the latent space are drawn from GMM:
\begin{align}
\bm{v} | \bm{z} 
\sim p(\bm{v} | \bm{z}; \eta, k), 
\quad 
\bm{z} 
\sim p(\bm{z}; \xi, k), 
\end{align}
where $p(\bm{v} | \bm{z}; \eta, k)$ 
and $p(\bm{z}; \xi, k)$ 
are  
the normal distribution and the categorical distribution, 
respectively, 
$\eta$ and $\xi$ are the parameters of the distributions, 
and $k$ is the number of mixture in GMM. 
%The Gaussian mixture model is a latent variable model 
%with a mixture of multiple normal distributions in \eqref{eq:original_probability_density}. 
%\begin{align}
%p(h^{0}; \{ \mu_{k} \}_{k=1}^{K}, \{ \Sigma_{k} \}_{k=1}^{K})
%= \sum_{k=1}^{K} w_{k} \mathcal{N}(h^{0}; \mu_{k}, \Sigma_{k}), 
%\quad \sum_{k=1}^{N} w_{k} = 1, \label{eq:original_probability_density}
%\end{align}
%where $\mu_{k} \in \mathbb{R}^{D}$ 
%and $\Sigma_{k} \in \mathbb{R}^{D \times D}$ 
%denote 
%the mean and variance-covariance matrix 
%of the $k$-th component, 
%respectively. 
We select the best $k$ as $K$ 
%based on the minimum description length~(MDL)  principle~\cite{Rissanen1978Automatica,Yamanishi2023Springer}. 
based on MDL  principle~\cite{Rissanen1978Automatica,Yamanishi2023Springer}. 
MDL asserts that the best model should be chosen as one that minimizes the total code-length for encoding a model and data given the model. 
It is suitable to employ MDL for determination of the number of clusters of hierarchical latent variable models, such as GMM.  
In practice, we choose $K$ 
among a given candidate set $\mathcal{K}$ 
and 
the embedding dimension $d$ 
among a given candidate set $\mathcal{D}$ 
to minimize the code-length for encoding a set of the node embedding vectors $\bm{v}$ 
and a set of the latent variables $\bm{z}$. 
Note that $\mathcal{K}$ and $\mathcal{D}$ are set by users. 
See the experiments in  Sec.~\ref{section:experiments} for an empirical setting. 
%with GMM, as follows: 
The criterion for choosing $K$ and $D$ is as follows: 
\begin{align}
(K, \, D) = \argmin_{(k, d) \in \mathcal{K} \times  \mathcal{D}} \, \, \,
%L(\bm{v} \in \mathbb{R}^{N \times d}, \, 
%  \bm{z} \in \{1, \dots, k\}^{N}; k), 
L(\bm{v}, \, \bm{z}; k), 
\label{eq:determination_of_the_number_of_clusters_with_dnml_code_length}
\end{align}
where $L$ denotes a code-length for encoding $\bm{v} \in \mathbb{R}^{N \times d}$ 
and $\bm{z} \in \{ 1, \dots, k \}^{N}$ 
with $k$ mixture of normal distributions. 
For the choice of $L$, 
the decomposed normalized maximum likelihood~(DNML) code-length~\cite{Wu2017KDD,Yamanishi2019DAMI} is often employed for model selection of latent variable models, 
with applications such as 
graph summarization~\cite{Fukushima2021ComplexNetworks},
hierarchical change detection in latent variable models~\cite{Fukushima2020ICDM}, 
balancing graph summarization and change detection~\cite{Fukushima2023ICDM}, 
and 
graph augmentation~\cite{Li2023ICDM}. 
DNML code-length is defined as follows~\cite{Wu2017KDD,Yamanishi2019DAMI}:
\begin{align}
L_{\mathrm{DNML}}(\bm{v}, \bm{z}; k)
&\mydef 
L_{\mathrm{NML}}(\bm{v} | \bm{z}; k)
+
L_{\mathrm{NML}}(\bm{z}; k), 
\end{align}
where 
%$\bm{z} = \{ z_{i} \}_{i=1}^{N} \in \{1, \dots, k \}^{N}$ denotes a random variable that indicates membership of each data point in the latent space, 
%and 
$L_{\mathrm{NML}}$ is 
the normalized maximum likelihood~(NML) code-length~\cite{Rissanen1978Automatica}:
\begin{align}
L_{\mathrm{NML}}(\bm{v} | \bm{z}; k)
&= -\log{p(\bm{v} | \bm{z}; 
%\hat{\eta}(\bm{h}, \bm{z}), k)} %\\
\hat{\eta}, k)} 
%&\quad 
    +\log{
        \sum_{\bm{v}', \bm{z}'}
            p(\bm{v}' | \bm{z}'; 
             \hat{\eta}; 
             k)
    }, 
\label{eq:nml_code_length_for_observed_variable} \\
L_{\mathrm{NML}}(\bm{z}; k) 
&= -\log{p(\bm{z}; 
   %\hat{\xi}(\bm{z}), k)}
   \hat{\xi})}
   +\log{
        \sum_{\bm{z}'}
            p(\bm{z}'; \hat{\xi}, k)
   }, 
\label{eq:nml_code_length_for_latent_variable}
\end{align}
where $\hat{\eta}=\hat{\eta}(\bm{v}, \bm{z})$ 
and 
$\hat{\xi}=\hat{\xi}(\bm{z})$ are the maximum likelihood estimator of $\eta$ and $\xi$, respectively. 
NML code-length is optimal in the sense that it achieves the Shtarkov's minimax regret~\cite{Rissanen2012Cambridge,Yamanishi2023Springer}. 
The latter terms in Eq.~\eqref{eq:nml_code_length_for_observed_variable} 
and 
Eq.~\eqref{eq:nml_code_length_for_latent_variable} 
are called the stochastic complexity, 
which measure the complexity of the models. 
It is known that the computational complexity for calculating the stochastic complexity is $O(N+k)$ for both the normal distribution~\cite{Yamanishi2019DAMI}  
and the categorical distribution~\cite{Kontkanen2007}. 
We denote the sets of the estimated means and variance-covariance matrices as 
$\{ \hat{\mu}_{k} \}_{k=1}^{K}$ and 
$\{ \hat{\Sigma}_{k} \}_{k=1}^{K}$, 
respectively. 

\subsubsection{Graph Decoder}
\label{subsubsection:graph_decoder}

We decode data points $\bm{v}$ into the original graph with a decoder. 
We define the probability of connection between nodes $i$ and $j$ of a new graph 
by calculating the sigmoid function of the inner product of $v_{i}$ and $v_{j}$ as follows:
\begin{align}
\sigma(v_{i}, v_{j})
&= \frac{1}{
    %\left(
    1 + \exp{\left(
    -\frac{
        \transpose{(v_{i})} v_{j}
    }{
        \| v_{i} \|
        \| v_{j} \|
    }
\right)}
%\right)
}, 
%&= 1/ \left(1 + \exp{\left(
%    -\frac{
%        \transpose{(v_{i})} v_{j}
%    }{
%        \| v_{i} \|
%        \| v_{j} \|
%    }
%\right)}
%\right), 
\label{eq:sigmoid_function_of_inner_product}
\end{align}
where $\| \cdot \|$ denotes $L^{2}$ norm. 
The loss function of graph decoder is defined as the binary cross-entropy as follows:
\begin{align}
%L_{\mathrm{dec}}
\ell
&= \sum_{i, j}
    %\left\{
    %\biggl\{
     -e_{ij} \log{ \sigma(
            v_{i}, 
            v_{j}
     )}
     -(1 - e_{ij}) \log{
        (
            1-\sigma (
                v_{i}, 
                v_{j}
            )
        )
     }, 
    %\right\}, 
    %\biggr\}, 
\label{eq:loss_decoder}
\end{align}
where $e_{ij} \in \{0, 1\}$ denotes the connection between the $i$th and $j$th nodes. 
%The total loss to be minimized is defined as the weighted sum of Eq.~\eqref{eq:parameter_optimization_problem_in_inverse_process}  and Eq.~\eqref{eq:loss_decoder} as follows:
%\begin{align}
%L &= L_{\mathrm{diffgen}} + \lambda L_{\mathrm{dec}}, 
%\quad 
%\lambda \in \mathbb{R}^{+}. 
%\label{eq:total_loss}
%\end{align}
%We use the GMM-based diffusion model for generating categorical data~\cite{Regol2023AAAI} in the diffusion and the reverse processes, which employs the autoregressive transformer. 
%We use the GMM-based diffusion model~\cite{Regol2023AAAI} 
%in the diffusion and the reverse processes, which employs the autoregressive transformer. 
%However, the loss functions are different between our setting and theirs. 
%In \cite{Regol2023AAAI}, 
%they consider the (negative) log-likelihood loss in Eq.~\eqref{eq:parameter_optimization_problem_in_inverse_process}, 
%whereas we also consider the reconstruction loss in Eq.~\eqref{eq:loss_decoder}. 
The training step is summarized in Algorithm~\ref{algorithm:learning_step}. 

\begin{algorithm}[tb]
%\setstretch{0.95}
\caption{Training Step} \label{algorithm:learning_step}
\begin{small}
\begin{algorithmic}[1]
\REQUIRE $G=(X, A)$: 
a static graph, 
where 
$X \in \mathbb{R}^{N \times F}$ 
and 
$A \in \mathbb{R}^{N \times N}$ 
denote feature matrix of nodes, 
weighted adjacency matrix, 
and labels of nodes, 
respectively. 
$\mathcal{K}$: a set of candidate number of mixtures of GMM in the latent space. 
$\mathcal{D}$: a set of candidate embedding dimensions. 
%$\{ \beta_{t} \}_{t=1}^{T}$: noise schedular, 
%$\lambda$: a penalty term.  
$\eta$: learning rate for the novelty condition. 
$\mathrm{epoch}_{\max}$: the maximum epoch
\ENSURE $K$: the optimal number of clusters, $\{ \hat{\mu}_{k} \}_{k=1}^{K}$: the estimated set of means of GMM, $\{ \hat{\Sigma}_{k} \}_{k=1}^{K}$: the estimated set of variance-covariance matrices of GMM. 
%$\hat{\theta}$: the estimated parameter in the reverse process in DDPM. 
\FOR {$\mathrm{epoch} = 1, \dots, \mathrm{epoch}_{\max}$}
\STATE Encode $G$ with VGAE and obtain a latent representation $\bm{v}$. 
\STATE Decode $\bm{v}$ with the decoder according to Eq.~\eqref{eq:sigmoid_function_of_inner_product} 
and obtain the reconstructed graph. 
\STATE Calculate the loss according to Eq.~\eqref{eq:loss_decoder}. 
\ENDFOR 
\STATE Embed each node in $G$ 
with a graph embedding algorithm and obtain its latent representation $\bm{v} = \{ v_{i} \}_{i=1}^{N}$. 
\STATE 
Fit GMM to $\bm{v}$ 
and estimate $K \in \mathcal{K}$ according to Eq.~\eqref{eq:determination_of_the_number_of_clusters_with_dnml_code_length}, and accordingly, $\{ \hat{\mu}_{k} \}_{k=1}^{K}$ and $\{ \hat{\Sigma}_{k} \}_{k=1}^{K}$. 
%\STATE Estimate $\hat{\theta}$ by minimizing Eq.~\eqref{eq:total_loss}. 
\end{algorithmic}
\end{small}
\end{algorithm}

%\subsection{Novelty Generation Step}
%\label{subsection:novelty_generation_step}
\subsection{Community Augmentation Step}
\label{subsection:community_augmentation_step}

%In the novelty generation step, 
In the community augmentation step, 
we add a new cluster in the latent space, 
%diffuse and denoise them with DDPM trained with Algorithm~\ref{algorithm:learning_step}, 
and decode them to generate a novel graph. 
Given the estimated means $\{ \hat{\mu}_{k} \}_{k=1}^{K}$ 
and variance-covariance matrices $\{ \hat{\Sigma}_{k} \}_{k=1}^{K}$, 
we are interested in the issue of how we should place new clusters. 
In the following, 
we use $\{ \mu_{k} \}_{k=1}^{K}$ and $\{ \Sigma_{k} \}_{k=1}^{K}$ 
instead of $\{ \hat{\mu}_{k} \}_{k=1}^{K}$ 
and 
$\{ \hat{\Sigma}_{k} \}_{k=1}^{K}$, 
respectively, 
to make the notations consistent. 

\subsubsection{Formulation of Novelty and Reliability Conditions}
\label{subsubsection:formulation_of_novelty_and_reliability_conditions}

First, we consider how to formulate mathematically the novelty condition in Eq.~\eqref{eq:condition_between_existing_and_newly_added_clusters} 
and the reliability condition in 
Eq.~\eqref{eq:condition_between_existing_and_newly_generated_probability_densities}. 
%Let us set the number of added clusters to $k_{\mathrm{new}} \in \mathbb{N}$.  
Let us assume that we add a cluster. 
The original and new probabilistic densities of the latent variable $v$ are  denoted in 
Eq.~\eqref{eq:original_probability_density} and Eq.~\eqref{eq:probability_density_with_newly_generated_clusters}, 
respectively. 
%We are concerned with how we should formulate the following two distance conditions; 
%the novelty index in 
%\eqref{eq:condition_between_existing_and_newly_added_clusters} 
%and 
%the reliability index in 
%\eqref{eq:condition_between_existing_and_newly_generated_probability_densities}. 

We can select any (quasi-)distance measure in Eq.~\eqref{eq:condition_between_existing_and_newly_added_clusters} and Eq.~\eqref{eq:condition_between_existing_and_newly_generated_probability_densities}, 
but we simply adopt the KL~(Kullback-Leibler) divergence in this study. 
Then, 
for Eq.~\eqref{eq:condition_between_existing_and_newly_added_clusters}, 
%the distance condition between the existing and newly added clusters is expressed as follows: 
the novelty condition is expressed as follows 
for $k=1, \dots, K$: 
\begin{align}
&d_{\mathrm{KL}}(
    \mathcal{N}(v; 
        \mu_{k}, 
        \Sigma_{k}),  \, 
    \mathcal{N}(v; 
        \mu_{K+1}, 
        \Sigma_{K+1})
) \\ 
&= \int 
     \mathcal{N}(v; 
        \mu_{k}, 
        \Sigma_{k})
     \log{
        \frac{
            \mathcal{N}(v;
                \mu_{k}, 
                \Sigma_{k})
        }{
            \mathcal{N}(v;
                \mu_{K+1}, 
                \Sigma_{K+1})
        }
     } \, \mathrm{d}v 
\geq \delta_{0},  %\, 
%(k=1, \dots, K), 
\label{eq:kl_distance_between_existing_and_newly_added_clusters}
\end{align}
where $d_{\mathrm{KL}}$ denotes 
KL divergence: 
$d_{\mathrm{KL}}(p, q) \mydef \int p(x) \log{\frac{p(x)}{q(x)}} \, \mathrm{d}x$. 

By contrast, 
for Eq.~\eqref{eq:condition_between_existing_and_newly_generated_probability_densities}, 
%the distance condition between the existing and newly generated probability densities is expressed by combining \eqref{eq:condition_between_existing_and_newly_generated_probability_densities} and 
the reliability condition is expressed by combining 
Eq.~\eqref{eq:original_probability_density} 
and 
Eq.~\eqref{eq:probability_density_with_newly_generated_clusters},  
%\eqref{eq:original_probability_density}, 
%and 
%\eqref{eq:probability_density_with_newly_generated_clusters},  
as follows:
\begin{align} 
&d_{\mathrm{KL}}(p_{\mathrm{orig}}, 
 p_{\mathrm{new}})  
= 
\int 
  p_{\mathrm{orig}}
   \log{
     \frac{
       p_{\mathrm{orig}}
     }{
       p_{\mathrm{new}}
     }
   } \, \mathrm{d}v  \\
&=
\int 
  \sum_{k=1}^{K} w_{k} 
  \mathcal{N}(v; 
    \mu_{k}, 
    \Sigma_{k})
  \log{
    \frac{
      \sum_{k=1}^{K} 
      w_{k} \mathcal{N}(v; 
        \mu_{k},
        \Sigma_{k})
    }{
      \sum_{k'=1}^{K+1} 
      w_{k'}' \mathcal{N}(v; \mu_{k'}, \Sigma_{k'})
    }
  } 
  \mathrm{d}v
\leq \delta_{1}. 
\label{eq:kl_distance_between_existing_and_newly_generated_probability_densities}
\end{align}
Our goal is to find a mean $\mu_{K+1}$ 
and 
variance-covariance matrix  
$\Sigma_{K+1}$ 
that satisfy Eq.~\eqref{eq:kl_distance_between_existing_and_newly_added_clusters} 
and 
Eq.~\eqref{eq:kl_distance_between_existing_and_newly_generated_probability_densities},  
given $\{ \mu_{k} \}_{k=1}^{K}$ 
and $\{ \Sigma_{k} \}_{k=1}^{K}$. 
KL divergence between two probability densities can be  approximated with the variational upper  bound~\cite{Hershey2007ICASSP,Durrieu2012ICASSP} as 
%KL divergence between two probability densities can be  approximated with the variational upper  bound~\cite{Hershey2007ICASSP,Durrieu2012ICASSP}. 
%See details for the variational upper bound in the supplementary material. 
%%Then, 
%%\eqref{eq:kl_distance_between_existing_and_newly_generated_probability_densities} is upper-bounded as follows:
\begin{align}
d_{\mathrm{KL}}(p_{\mathrm{orig}}, p_{\mathrm{new}}) 
&\leq 
d_{\phi, \psi}(p_{\mathrm{orig}}, p_{\mathrm{new}}). 
\label{eq:upperbound_of_distance_between_original_and_new_probabilistic_distributions}
\end{align}
See Sec.~\ref{section:detailed_derivation} in Appendix for details of the formulation of $d_{\phi, \psi}(p_{\mathrm{orig}}, p_{\mathrm{new}})$. 

\subsubsection{Numerical Methods for Novelty and Reliability Conditions}
\label{subsubsection:numerical_methods_for_novelty_and_reliability_conditions}

Then, we consider how to solve numerically  Eq.~\eqref{eq:kl_distance_between_existing_and_newly_added_clusters} for the novelty condition, 
and Eq.~\eqref{eq:inequality_upperbound_of_distance_between_original_and_new_probabilistic_distributions} 
for the reliability condition. 

For the novelty condition, 
Eq.~\eqref{eq:kl_distance_between_existing_and_newly_added_clusters} is expanded as 
\begin{align}
&d_{\mathrm{KL}}(
    \mathcal{N}(v; \mu_{k}, \Sigma_{k}), \, 
    \mathcal{N}(v; \mu_{K+1}, 
    \Sigma_{K+1})
)  \\
&= 
\frac{1}{2} 
\Bigl\{
    \transpose{(\mu_{K+1} -\mu_{k})}
    \Sigma_{K+1}^{-1}
    (\mu_{K+1} - \mu_{k})
    +\mathrm{tr}(\Sigma_{K+1}^{-1} \Sigma_{k}) \\
&\quad \quad 
    -\log{ 
       \frac{
         |\Sigma_{K+1}|
       }{
         |\Sigma_{k}|
       } 
     }
     -D
\Bigr\}. 
\label{eq:kl_divergence_of_normal_distributions}
\end{align}
We update $\mu_{K+1}$ and $\Sigma_{K+1}$ with the stochastic gradient ascent~(SGA), 
where the gradients are $\frac{\partial d_{\mathrm{KL}}}{\partial \mu_{K+1}}$ and $\frac{\partial d_{\mathrm{KL}}}{\partial \Sigma_{K+1}}$. 
The update rule for $\mu_{K+1}$ is straightforward because the gradient $\frac{\partial d_{\mathrm{KL}}}{\partial \mu_{K+1}}$ is relatively easy to derive theoretically and to calculate computationally, 
while the update rule for $\Sigma_{K+1}$ is not straightforward. 
%We provide the derivation of the update rule in Sec.~\ref{subsection:derivation_of_Von_Neumann_Operator_for_Mirror_Descent_Algorithm} in Appendix. 
We derive the update rule 
with von Neumann operator for mirror descent algorithm, 
in reference to \cite{Tsuda2005JMLR}. 
The update rule for $\Sigma_{K+1}$ at the $j$th step is derived as 
\begin{align}
\Sigma_{K+1}^{(j+1)} 
&= \argmin_{\Sigma} 
   \left\{
    \Delta_{F}(
        \Sigma, 
        \Sigma_{K+1}^{(j)}
    )
    +
    \eta \nabla L(\Sigma_{K+1}^{(j)})
   \right\}, 
\label{eq:estimated_variance-covariance_matrix_with_von_Neumann_operator}
\end{align}
where $\eta$ is a learning rate and $\Delta_{F}(\Sigma, \, \Sigma_{K+1}^{(j)})$ is defined as 
\begin{align}
\Delta_{F}(\Sigma, \, \Sigma_{K+1}^{(j)})
&= \mathrm{tr} 
   \left(
     \Sigma \log{\Sigma} 
     -
     \Sigma_{K+1}^{(j)} \log{\Sigma_{K+1}^{(j)}}
   \right). 
\label{eq:delta_F}
\end{align}
The logarithm of a matrix in Eq.~\eqref{eq:delta_F}
is defined as 
$\log{\Sigma} = V (\log{\Lambda}) \transpose{V}$ 
with the singular value decomposition~(SVD) of $\Sigma$: 
$\Sigma = V \Lambda \transpose{V}$, 
where $\Lambda \in \mathbb{R}^{D \times D}$ is a diagonal matrix with the singular values, 
and $V \in \mathbb{R}^{D \times D}$ is a unitary matrix.  
By substituting Eq.~\eqref{eq:delta_F} into Eq.~\eqref{eq:estimated_variance-covariance_matrix_with_von_Neumann_operator}, 
we obtain the following update rule:
\begin{align}
\Sigma_{K+1}^{(j+1)} 
&= \exp{
    \left(
        \log{
          \Sigma_{K+1}^{(j)}
        }
        +
        \eta \nabla L(\Sigma_{K+1}^{(j)})
    \right)
},  
\end{align}
where the exponential operator for a matrix is defined as 
$\exp{\Sigma} = V (\exp{\Lambda}) \transpose{V}$. 
This procedure is repeated until Eq.~\eqref{eq:kl_distance_between_existing_and_newly_added_clusters} is satisfied. 
There are many ways to set the initial values of $\mu_{K+1}$ and $\Sigma_{K+1}$. 
In this study, we set them to ones of the existing clusters. 
However, this is not the only solution, thus, it remains one of future work to address this issue. 

By contrast, for the reliability condition, 
we consider how to numerically %estimate $\phi_{k'|k}$ and $\psi_{k|k'}$ for $k=1, \dots, K$ and $k'=1, \dots, K+1$. 
minimize $d_{\phi, \psi}(p_{\mathrm{orig}}, \, p_{\mathrm{new}})$.  
%in Eq.~\eqref{eq:upperbound_of_distance_between_original_and_new_probabilistic_distributions}.  
%See Appendix for details. 
%We iteratively optimize $\psi_{k|k'}$ for fixed $\phi_{k|k'}$, 
%and optimize 
%$\phi_{k|k'}$ for fixed $\psi_{k|k'}$~\cite{Hershey2007ICASSP,Durrieu2012ICASSP}. 
Meanwhile, 
we also update $\{ w_{k'}' \}_{k'=1}^{K+1}$. 
It is necessary to satisfy $\sum_{k'=1}^{K+1} w_{k'}' = 1$ when we update the weights of clusters. 
Therefore, we adopt the mirror descent exponential gradient algorithm~\cite{Kivinen1997IC} for fixed $\{ \mu_{k} \}_{k=1}^{K+1}$,  
$\{ \Sigma_{k} \}_{k=1}^{K+1}$, 
$\hat{\phi}$, and $\hat{\psi}$. 
In the $j$th step, $w_{k'}'$ is updated as follows: %as follows:
\begin{align}
%\frac{\partial d_{\phi, \psi}}{\partial w_{k}}(p_{\mathrm{orig}}, p_{\mathrm{new}})
%&= - \frac{\partial}{\partial w_{k}}
%    \sum_{k, k'} 
%        \hat{\phi}_{k'}
%        \int 
%            \mathcal{N}(h; \mu_{k}, \Sigma_{k})
w_{k'}'{}^{(j+1)} 
&=  \frac{ 
        w_{k'}'{}^{(j)}
        \exp{(-\eta g_{k'}^{(j)})}
    }{
        \sum_{\ell=1}^{K+1} 
        w_{\ell}'^{(j)}
        \exp{(-\eta g_{\ell}^{(j)})}    
    }, 
%\quad 
\label{eq:mirror_descent_exponential_gradient_algorithm}
\end{align}
for $k'=1, \dots, K+1$, 
where 
$g_{k'}^{(j)}
= \frac{\partial d_{\phi, \psi}}{\partial w_{k'}'{}^{(j)}}$ 
is the gradient of KL divergence in Eq.~\eqref{eq:kl_distance_between_existing_and_newly_generated_probability_densities} at the $j$th step with respect to $w'_{k}$, 
and $\eta \in \mathbb{R}^{+}$ is a learning rate. 
We repeat this procedure until the following inequality holds:
\begin{align}
d_{\hat{\phi}, \hat{\psi}}(p_{\mathrm{orig}}, 
 p_{\mathrm{new}})
&\leq \delta_{1}. 
\label{eq:inequality_upperbound_of_distance_between_original_and_new_probabilistic_distributions}
\end{align}
%Thus, we obtain 
%$\mu_{K+1}$ 
%and 
%$\Sigma_{K+1}$ 
%by combining \eqref{eq:kl_distance_between_existing_and_newly_added_clusters} 
%and 
%\eqref{eq:inequality_upperbound_of_distance_between_original_and_new_probabilistic_distributions}. 
%Derivatives of KL divergence in 
%\eqref{eq:kl_divergence_of_normal_distributions} 
%with respect to $\mu_{K+1}$ and $\Sigma_{K+1}$ 
%are expanded as follows: 
%\begin{align}
%\frac{
%  \partial d_{\mathrm{KL}}
%}{
%  \partial \mu_{K+1}
%}
%(\mathcal{N}(\mu_{k}, 
% \Sigma_{k}),
% \mathcal{N}(\mu_{K+1}, 
% \Sigma_{K+1})
%)
%&= \Sigma_{K+1}^{-1}
%    (\mu_{K+1} - \mu_{k}), 
%\label{eq:partial_derivative_of_KL_wrt_mean} 
%\\
%\frac{
%    \partial d_{\mathrm{KL}}
%}{
%    \partial \Sigma_{K+1}
%}(\mathcal{N}(\mu_{k}, 
%  \Sigma_{k}), 
%  \mathcal{N}(\mu_{K+1}, 
%  \Sigma_{K+1})
% )
%&= 
%\label{eq:partial_derivative_of_KL_wrt_cov_matrix}
%\end{align}
%We iteratively update $\mu_{K+1}$ and $\Sigma_{k+1}$ 
%with stochastic gradient ascent~(SGA) 
%until \eqref{eq:kl_distance_between_existing_and_newly_added_clusters} holds. 
When Eq.~\eqref{eq:inequality_upperbound_of_distance_between_original_and_new_probabilistic_distributions} is satisfied, 
we obtain the estimators 
$\{ \hat{w}_{k'} \}_{k'=1}^{K+1}$. 
%We repeat this procedure until \eqref{eq:inequality_upperbound_of_distance_between_original_and_new_probabilistic_distributions} is satisfied. 

Then, we generate data points $\bm{v}' \in \mathbb{R}^{M \times D}$ 
by sampling $M \in \mathbb{N}$ data points from $\mathcal{N}(v; \mu_{K+1}, \Sigma_{K+1})$. 
%and concatenate the data points into $\bm{h}'^{0} \in \mathbb{R}^{M \times D}$, 
%where $M = \sum_{k=K+1}^{K+k_{\mathrm{new}}} M_{k}$. 
%We diffuse $\bm{h}'^{0}$ in the diffusion process and obtain $\bm{h}'^{T} \in \mathbb{R}^{M \times D}$. 
%We then denoise $\bm{h}'^{T}$ iteratively with $\hat{\theta}$ and obtain $\bar{\bm{h}}'^{0} \in \mathbb{R}^{M \times D}$. 
We concatenate $\bm{v}$ and $\bm{v}'$ 
and the obtained matrix is represented as 
%$\bm{v} \oplus \bm{v}' \in \mathbb{R}^{(N+M) \times D}$,  
$\bm{v} \| \bm{v}' \in \mathbb{R}^{(N+M) \times D}$,  
and then estimate the number of clusters $K_{\mathrm{est}}$ according to the DNML code-length criteria in  Eq.~\eqref{eq:determination_of_the_number_of_clusters_with_dnml_code_length}. 
If $K_{\mathrm{est}} \neq K+1$, 
we repeat this procedure again 
because the obtained cluster is not satisfactory. 
When $K_{\mathrm{est}} = K+1$, 
we then estimate the probability of connection between nodes 
with Eq.~\eqref{eq:sigmoid_function_of_inner_product}. 
In that way, we obtain a new graph $G'$. 

In practice, we have to determine the threshold for the decoder in applying the sigmoid function in Eq.~\eqref{eq:loss_decoder}. 
In this study, 
we set the threshold to the average scores of the existent edges in the training dataset~\cite{Li2023ICDM}. 
The threshold values in the novelty and reliability conditions, 
that is, 
$\delta_{0}$ and $\delta_{1}$ are also hyperparameters. 
In the experiments in Sec.~\ref{section:experiments}, 
we set $\delta_{0}$ to 
%five times 
the maximum KL divergence between the existing clusters. 
Note that it is more preferable to specify a range of $\delta_{0}$ and the one of $\delta_{1}$ than to specify a value of $\delta_{0}$ and the one of $\delta_{1}$ in practice. 
We examine the sensitivity of the performance of GCA through extensive experiments in Sec.~\ref{subsection:sensitivity_analysis_on_hyperperamters}, 
and support this suggestion. 
This is because we would like to generate and select novel graphs with a new community structure. 
%The novelty generation step is summarized in Algorithm~\ref{algorithm:novelty_generation_step}. 
%The novelty generation step is summarized in Algorithm~\ref{algorithm:community_augmentation_step}. 
The community augmentation step is summarized in Algorithm~\ref{algorithm:community_augmentation_step}. 
%Note that the parameters of the diffusion model such as a set of noise schedulars $\{ \beta_{t} \}_{t=1}^{T}$ are omitted in the inputs of Algorithm~\ref{algorithm:novelty_generation_step}. 

\subsection{Computational Complexity}
\label{subsection:computational_complexity}

For the training step, 
%the time complexity is largely influenced by VGAE, GMM, DNML code-length calculation, and diffusion model. 
the time complexity is largely influenced by VGAE, GMM, and DNML code-length calculation. 
The time complexity of VGAE per iteration is $O(|E|)$, where $E$ denotes the set of edges in the original graph. 
The time complexity of GMM per iteration is $O(NKD^{2})$. 
%The time complexity of the diffusion model is $O(T^{2})$, 
%where $T$ is the number of steps in the diffusion model. 
Therefore, the total time complexity in training step is $O(I_{1} |E| + I_{2} NKD^{2})$, 
where $I_{1}$ and $I_{2}$ are the numbers of iterations required for VGAE and GMM, respectively. 
%In summary, the time complexity in the training step is $O(I_{1} |E| + I_{2}NKD^{2} + I_{3} T^{2})$. 
Note that the time complexity is  linear with respect to the numbers of edges and nodes in the original graph, 
and the number of clusters in the latent space.  

%For the novelty generation step, 
For the community augmentation step, 
the time complexity is largely influenced by DNML code-length calculation, SGA, and mirror descent exponential gradient algorithm. 
DNML code-length calculation requires $O(N+K)$, 
SGA requires $O(I_{3} b_{1}(K+1))$, 
and 
mirror descent exponential gradient requires $O(I_{4} b_{2} (K+1))$, 
where $b_{1}$ and $b_{2}$ are batch sizes in SGA and mirror descent exponential gradient, respectively. 
In summary, 
%the time complexity of the novelty generation step is $O(N+(I_{3}b_{1} + I_{4}b_{2})K)$. 
the time complexity of the community augmentation step is $O(N+(I_{3}b_{1} + I_{4}b_{2})K)$. 
Note that this time complexity is linear with respect to $N$ and $K$.

\begin{algorithm}[tb]
%\setstretch{0.95}
%\caption{Novelty Generation Step} \label{algorithm:novelty_generation_step}
\caption{Community Augmentation Step} \label{algorithm:community_augmentation_step}
\begin{small}
\begin{algorithmic}[1]
\REQUIRE 
$\bm{v}$: a latent representation. 
%$\bar{\bm{v}}$: a latent representation. 
%$T$: the number of steps in the  diffusion model. 
$\{ \hat{\mu}_{k} \}_{k=1}^{K}$: estimated means of clusters. 
$\{ \hat{\Sigma}_{k} \}_{k=1}^{K}$: estimated variance-covariance matrices of clusters. 
%$\hat{\theta}$: estimated parameter. 
$\delta_{0}$: a distance threshold parameter 
between the original and newly-generated probability density,  
$\delta_{1}$: a distance threshold parameter between each pair of clusters in the original and newly-added clusters. 
%$k_{\mathrm{new}}$: the number of newly-added clusters. 
%$\{ M_{K+1}, \dots, M_{K+k_{\mathrm{new}}} \}$: a set of the numbers of generated nodes in the new graph. 
$M$: the number of generated nodes in the new graph. 
$\mathcal{K}$: a set of candidate number of mixtures of GMM in the latent space. 
\ENSURE a new graph $G'$. 
%\STATE Initialize $\{ \mu_{k} \}_{k=K+1}^{K+k_{\mathrm{new}}}$ 
\STATE Initialize $\mu_{K+1}$,  
$\Sigma_{K+1}$, 
and $\{ w_{k'}' \}_{k'=1}^{K+1}$ \, ($\sum_{k'=1}^{K+1} w_{k'}' = 1$). 
\WHILE {Eq.~\eqref{eq:inequality_upperbound_of_distance_between_original_and_new_probabilistic_distributions} is not satisfied}
  \STATE update $\mu_{K+1}$, $\Sigma_{K+1}$ with SGA. 
  \STATE update $\{ w_{k'}'  \}_{k'=1}^{K+1}$ with the mirror descent exponential gradient algorithm according to Eq.~\eqref{eq:mirror_descent_exponential_gradient_algorithm}. 
\ENDWHILE
%\STATE Sample $M_{k}$ data points from $\mathcal{N}(\mu_{k}, \Sigma_{k})$ \, $(k=K+1, \dots, K+k_{\mathrm{new}})$. 
    %\STATE Generate $\bm{h}'^{0}$ by concatenating the sampled data points. 
\WHILE{$K_{\mathrm{est}} \neq K+1$}
    \STATE Generate $\bm{v}'$ by sampling $M$ data points from $\mathcal{N}(\bm{v}; \mu_{K+1}, \Sigma_{K+1})$ 
    %\STATE Obtain $\bm{v}'^{T}$ by diffusing $\bm{h}'^{0}$ 
    %according to Eq.~\eqref{eq:diffusion_process_one_step}. 
    %\STATE Denoise $\bm{v}'$ with $\hat{\theta}$ 
    and obtain $\bm{v}'$. 
    \STATE Estimate $K_{\mathrm{est}} = \argmin_{k' \in \mathcal{K}} L(\bm{v} \| \bm{v}'; k')$. 
\ENDWHILE
\STATE Decode $\bm{v} \oplus \bm{v}'$ and obtain a new graph $G'$. 
\end{algorithmic}
\end{small}
\end{algorithm}

\section{Experiments}
\label{section:experiments}

In this section, 
we demonstrate the effectiveness of the proposed algorithm on both synthetic and real datasets. 
The source code is provided at \url{https://github.com/s-fuku/gca}. 
We used Ubuntu~20.04 and NVIDIA A100 80GB$\times$2. 
The research questions to be answered through the experiments are summarized as follows:
\begin{description}
\item[RQ1.] How well does  GCA  work compared to existing algorithms in graph generative models?
%\item[RQ2.] How is NGG-DM affected by the hyperparameters $\delta_{0}$ and $\delta_{1}$?
\item[RQ2.] How does GCA depend on the hyperparameters? 
\item[RQ3.] How well does each part of GCA work?  Can GCA be improved  by replacing it with other techniques?
%\item[RQ4.] How does GCA work better if we refine each part of GCA?
\end{description}
%We answer the research questions in the following experiments. 

\subsection{Synthetic Dataset~(RQ1)}
\label{subsection:synthetic_dataset}

\subsubsection{Data}
\label{section:synthetic_dataset_data}

%We used the following two datasets~\cite{Martinkus2022ICML,Jo2023arXiv}: 
We used SBM  dataset~\cite{Martinkus2022ICML,Jo2023arXiv}~\footnote{\url{https://drive.google.com/drive/folders/1imzwi4a0cpVvE_Vyiwl7JCtkr13hv9Da}}. 
This dataset is composed of 200 synthetic stochastic block model~\cite{Snijders1997JoC} graphs with the number of communities uniformly sampled between two and five, 
where the number of nodes in each community is uniformly sampled between 20 and 40. 
Each node in each graph has two features; 
eigenvalues of the graph 
laplacians. 
%\begin{itemize}
%\item \textbf{SBM}\footnote{\url{https://drive.google.com/drive/folders/1imzwi4a0cpVvE_Vyiwl7JCtkr13hv9Da}}: 200 synthetic stochastic block model~\cite{Snijders1997JoC} graphs with the number of communities uniformly sampled between two and five, 
%where the number of nodes in each community is uniformly sampled between 20 and 40. 
%Each node in each graph has two features; eigenvalues of the graph laplacians. 
%
%\item \textbf{Planar}\footnote{\url{https://drive.google.com/file/d/1B1RQFomDxHKnFlEBQnNsF8rbZ7tt-4RA}}: 200 synthetic planar graphs 
%where each graph has 64 nodes. 
%For this dataset, 
%each node in each graph has the two features as in SBM  dataset. 
%\end{itemize}
%For each dataset, 
The numbers of graphs in training dataset, validation dataset, and test dataset are 128, 32, and 40, respectively. 
%We trained VGAE and GMM-based diffusion model with the training dataset in it, 
We trained VGAE with the training dataset, 
and evaluated their performance with the test dataset. 
%Note that we trained each graph in the training dataset and then generated graphs with a new community structure. 
Note that we generated graphs with a new community structure using the test dataset.

\subsubsection{Methods for Comparison}
\label{subsubsection:methods_for_comparison}

We chose the following three methods for comparison; 
GDSS\footnote{\url{https://github.com/harryjo97/GDSS}}\cite{Jo2022ICML}, 
and 
ConGress 
and DiGress\footnote{\url{https://github.com/cvignac/DiGress}}\cite{Vignac2023ICLR}. 
%They are known to be the state-of-the-art graph generative models with diffusion models. 
They are known to be the state-of-the-art graph generative models with diffusion models~\cite{Ho2020NeurIPS}. 
%The detailed implementations 
%and experimental settings 
%of GDSS, ConGress, and DiGress are summarized in Appendix. 

\subsubsection{Evaluation Metrics}
\label{subsubsection:evaluation_metrics}

We evaluated the performance of GCA and the rival algorithms %in terms of 
from the following two perspectives; 
(i) the degree of novelty generation, 
and 
(ii) the degree of maintenance of structure of original graphs. 
We generated 100 graphs for evaluation. 

First, for (i), 
we employed the evaluation setting in previous studies~\cite{Martinkus2022ICML,Jo2023arXiv}; the percentage of valid, unique, and novel~(V.U.N.). 
V.U.N. has been employed for evaluating how well the trained generative models work for generating graphs with semantically and grammatically meaningful structure, less duplication, and nonexistent in the graph dataset for evaluation. 
However, V.U.N. might not be proper to measure \textit{novelty} in some cases. 
Therefore, in addition to GCA, 
we introduce an anomaly score~(Anomaly) as the log-probability of the generated graphs for GCA: 
%Anomaly is defined as the logarithm of the probability densities of the latent variables 
%For NGG-DM, Anomaly is defined as follows: $\mathrm{Anomaly} = -\log{p_\mathrm{orig}(\bm{h})}$. 
$\mathrm{Anomaly} = -\log{p_{\mathrm{orig}}(v)}$, 
where $v \in \mathbb{R}^{D}$ is an embedding vector in a latent space 
and $p_{\mathrm{orig}}$ is the original probabilistic density function in Eq.~\eqref{eq:original_probability_density}. 
For GDSS, ConGress, and DiGress, 
we defined the Anomaly as 
the sum of the negative log-likelihood of the probability density functions of the generated node feature matrix and adjacency matrix. 
See the original paper of GDSS~\cite{Jo2022ICML}, 
and Congress and DiGress~\cite{Vignac2023ICLR} for the details of the probability density functions. 
%See the details for Appendix. 

Next, for (ii), 
we followed the evaluation setting in previous studies~\cite{Martinkus2022ICML,Jo2023arXiv}. 
We employed the maximum mean discrepancy~(MMD) of the following four graph statistics 
between the set of generated graphs and the test set; 
degree~(Deg.), 
cluster coefficient~(Clus.), 
count of orbits with 4 nodes~(Orbit), 
and the eigenvalues of the graph Laplacians~(Spec.). 
We followed \cite{Jo2023arXiv} for definitions of metrics 
with more comprehensive experiments than the original paper of DiGress~\cite{Vignac2023ICLR}. 

\subsubsection{Hyperparameter Setting}
\label{subsubsection:hyperparameter_setting_synthetic_dataset}

We set the candidate set of the number of clusters 
$\mathcal{K} = \{2, \dots, 10\}$, 
and the candidate set of the embedding dimension 
$\mathcal{D} = \{ 4, 8, 16, 24, 32 \}$ 
%for %both 
%SBM and Planar datasets, 
and selected the best dimension that minimizes the DNML code-length criterion in Eq.~\eqref{eq:determination_of_the_number_of_clusters_with_dnml_code_length}. 
%for each graph in the training dataset. 
We randomly set the initial values of the newly-added cluster, that is, 
the mean $\mu_{K+1}$ and the variance-covariance matrix $\Sigma_{K+1}$, to ones of the existing clusters; $\{ \mu_{k} \}_{k=1}^{K}$ and $\{ \Sigma_{k} \}_{k=1}^{K}$. 
The set of generated nodes are  set to $\mathcal{M} = \{ 5$, $10$, $15$, $20$, $25\}$. 
We set $\delta_{0}$ to $5$, 
$\delta_{1}$ to 
%five times of 
the maximum KL distance between the existing clusters~($= D_{\max}$). 
%and $\lambda=0.1$. 
%The detailed parameter setting, 
%in particular, 
%the configurations of parameters of the graph autoencoder and the diffusion model are described in the supplementary material. 

For the architecture of GVAE, 
we set the size of the hidden layer to the twice size of $X$~($=F$) 
and activation function to ReLU, in reference to \cite{Kipf2016arXiv}. 
In training VGAE, 
we set the number of epochs to 200 
%batch size to 16, 
and optimizer to Adam~(learning rate$=0.01$, and other parameters were set as the default values in PyTorch). 
In the community augmentation step, 
we set the learning rate for the novelty condition $\eta=0.01$. 

\subsubsection{Results}

%We changed the number of nodes added in the new component, denoted as $N$ among $N \in \{10$, $50$, $100$, $200$, $500\}$, 
%and then compared its performance with ones of the compared algorithms. 

%Table~\ref{table:evaluation_results_on_the_synthetic_datasets} shows the evaluation results on the synthetic datasets. 
Table~\ref{table:evaluation_results_on_the_synthetic_datasets} shows the evaluation results on the synthetic dataset. 
%Note that we trained each graph in the training dataset and generated graphs with the trained model. 
Note that we generated each graph with the test dataset and the trained model. 
%We selected $(K, D) = (2, 8)$ for Planar dataset, 
%$(K, D) = (2, 8)$, $(3, 16)$, $(4, 16)$, and $(5, 16)$ for SBM dataset. 
%Note that there are SBM graphs with different numbers of clusters between $2$ and $5$, 
%as we described it in Sec.~\ref{section:synthetic_dataset_data}. 
%For each pair of $K$ and $D$, we trained GCA separately. 
%The numbers of graphs contained in the estimated $K$ are $52$~($K=2$), $44$~($K=3$), $43$~($K=4$), and $61$~($K=5$). 
We observed that
GCA generated more novel graphs % out of the probability distributions 
in terms of V.U.N. and Anomaly. 
By contrast, 
GCA was comparable to the other algorithms in keeping structure of graphs in terms of  Deg., Clus., Orbit, and Spec. 
%on both SBM and Planar datasets. 
It should be noted that 
the properties of the generated graphs with GCA did not change greatly. 
It indicates that GCA keeps the structure of the original graph up to a number of added nodes in the new cluster. 

\begin{table*}[tb]
\caption{Evaluation results on the synthetic datasets. 
Time(T) and Time(A) indicate the computational time for training and community augmentation, respectively. 
``Training dataset'' indicates the MMD statistics between training and test datasets. The bold faces and the underlines mean the best and the second best scores, respectively. }
\label{table:evaluation_results_on_the_synthetic_datasets}
\centering
\begin{footnotesize}
{\tabcolsep=0.6\tabcolsep
\begin{tabular}{lrrrrrrrrrrrrrrrr}
\toprule
%   &
%\multicolumn{7}{c}{SBM} & 
%\multicolumn{7}{c}{Planar} \\
%\midrule
\multicolumn{1}{l}{} & 
\multicolumn{1}{c}{V.U.N.} & 
\multicolumn{1}{c}{Anomaly} & 
\multicolumn{1}{c}{Deg.} & 
\multicolumn{1}{c}{Clus.} & 
\multicolumn{1}{c}{Orbit} & 
\multicolumn{1}{c}{Spec.} & 
\multicolumn{1}{c}{Time(T)} &
\multicolumn{1}{c}{Time(A)} & 
%\multicolumn{1}{c}{V.U.N.} & 
%\multicolumn{1}{c}{Anomaly} & 
%\multicolumn{1}{c}{Deg.} & 
%\multicolumn{1}{c}{Clus.} & 
%\multicolumn{1}{c}{Orbit} & 
%\multicolumn{1}{c}{Spec.} & 
%\multicolumn{1}{c}{Time(T)} &
%\multicolumn{1}{c}{Time(A)} 
\\
\midrule
Training dataset & 
$100.0$ & 
\multicolumn{1}{c}{---} & 
$0.0008$ & 
$0.0322$ & 
$0.0255$ & 
$0.0063$ & 
\multicolumn{1}{c}{---} & 
\multicolumn{1}{c}{---} & % SBM 
%$100.0$ & 
%\multicolumn{1}{c}{---} & 
%$0.0002$ & 
%$0.0310$ & 
%$0.0005$ & 
%$0.0052$ & 
%\multicolumn{1}{c}{---} & 
%\multicolumn{1}{c}{---} % Planar 
\\
\midrule
%SPECTRE & $0.0005$ & $0.0785$ & $0.0012$ & $0.0112$ & $25$ & $0.61 \pm 0.12$& \\
%ERGMS &
%&
%&
%&
%&
%&
%&
%&   % Planar
%&
%&
%&
%&
%&
%&
%&   % SBM
%\\
%GPT-GNN &
%&
%&
%&
%&
%&
%&
%&   % Planar
%&
%&
%&
%&
%&
%&
%&   % SBM
%\\
GDSS & 
$5.0$ &
$0.65 \pm 0.09$ & 
$0.0212$ & 
$0.0646$ & 
$0.0894$ & 
$\mathbf{0.0128}$ &
177m & 3m20s 
%& % SBM 
%$0.0$ & 
%$0.64 \pm 0.13$ & 
%$0.0041$ & 
%$0.2676$ & 
%$0.1720$ & 
%$0.0370$ & 
%188m & 3m24s
% Planar 
\\
ConGress & 
$0.0$ & 
$0.41 \pm 0.15$ &
$0.0273$ & 
$0.1029$ & 
$0.1148$ &
$\underline{0.0132}$& 
162m & 3m26s 
%& % SBM
%$0.0$ & 
%$0.68 \pm 0.08$ & 
%$0.0048$ & 
%$0.2728$ & 
%$1.2950$ & 
%$0.0418$ & 
%161m & 3m18s % Planar 
\\
DiGress &
$75.0$ & 
$0.28 \pm 0.08$ & 
$\mathbf{0.0003}$ & 
$\mathbf{0.0372}$ & 
$\mathbf{0.0434}$ & 
$0.0400$ & 
$\underline{158}$m & % SBM 
3m08s %& 
%$75.0$ & 
%$0.39 \pm 0.15$ & 
%$\mathbf{0.0003}$ & 
%$\mathbf{0.0372}$ & 
%$\mathbf{0.0015}$ & 
%$\mathbf{0.0106}$ & 
%$\underline{153}$m & 
%\underline{3m15s}  % Planar 
\\
GCA~($M=5$) & 
$\mathbf{85.0}$ & 
$5.78 \pm 0.74$ & 
$\underline{0.0009}$ & 
$\underline{0.0587}$ & 
$\underline{0.0732}$ & 
$0.0412$ & 
$\mathbf{19m}$ & 
$\mathbf{20s}$ %& % SBM
%$\mathbf{79.0}$ & 
%$3.28 \pm 0.85$ &
%$\underline{0.0005}$ & 
%$\underline{0.0813}$ & 
%$\underline{0.0019}$ & 
%$\underline{0.0210}$ & 
%$\mathbf{17m}$ & $\mathbf{8s}$  % Planar 
\\
GCA~($M=10$) & 
$\underline{82.0}$ & 
$5.63 \pm 0.93$ & 
$0.0027$ & 
$0.0734$ & 
$0.0787$ & 
$0.0415$ & 
$\mathbf{19m}$ &  
\underline{22s} %& % SBM 
%$\underline{77.0}$ & 
%$3.93 \pm 0.93$ &
%$0.0021$ & 
%$0.1132$ & 
%$0.0023$ & 
%$0.0305$ & 
%$\mathbf{17m}$ &  
%$9$s % Planar 
\\  
GCA~($M=15$) & 
$79.0$ & 
$\mathbf{6.13 \pm 0.69}$ & 
$0.0035$ & 
$0.0831$ & 
$0.0898$ & 
$0.0419$ & 
$\mathbf{19m}$ &  
\underline{22s} %& % SBM 
%$73.0$ & 
%$\mathbf{4.20 \pm 0.84}$ & 
%$0.0023$ & 
%$0.1329$ & 
%$0.0029$ & 
%$0.0398$ & 
%$\mathbf{17m}$ &  
%$9$s % Planar 
\\  
GCA~($M=20$) & 
$75.0$ & 
$5.96 \pm 0.65$ & 
$0.0047$ & 
$0.0922$ & 
$0.1055$ & 
$0.0425$ & 
$\mathbf{19m}$ &  % SBM 
$32$s %&
%$71.0$ & 
%$3.95 \pm 1.15$ & 
%$0.0026$ & 
%$0.1653$ & 
%$0.0035$ & 
%$0.0478$ & 
%$\mathbf{17m}$ &  
%$10$s  % Planar 
\\  
GCA~($M=25$) & 
$72.0$ & 
$\underline{5.88 \pm 0.51}$ & 
$0.0073$ & 
$0.1021$ & 
$0.1092$ & 
$0.0435$ & 
$\mathbf{19m}$ &
$38$s %&  % SBM 
%$66.0$ & 
%$\underline{4.10 \pm 0.92}$ & 
%$0.0029$ & 
%$0.2149$ & 
%$0.0043$ & 
%$0.0650$ & 
%$\mathbf{17m}$ &  
%$12$s %Planar  
\\ 
\bottomrule
\end{tabular}
}
\end{footnotesize}
\end{table*}

\begin{figure}[tb]
\centering
\includegraphics[width=0.5\linewidth]{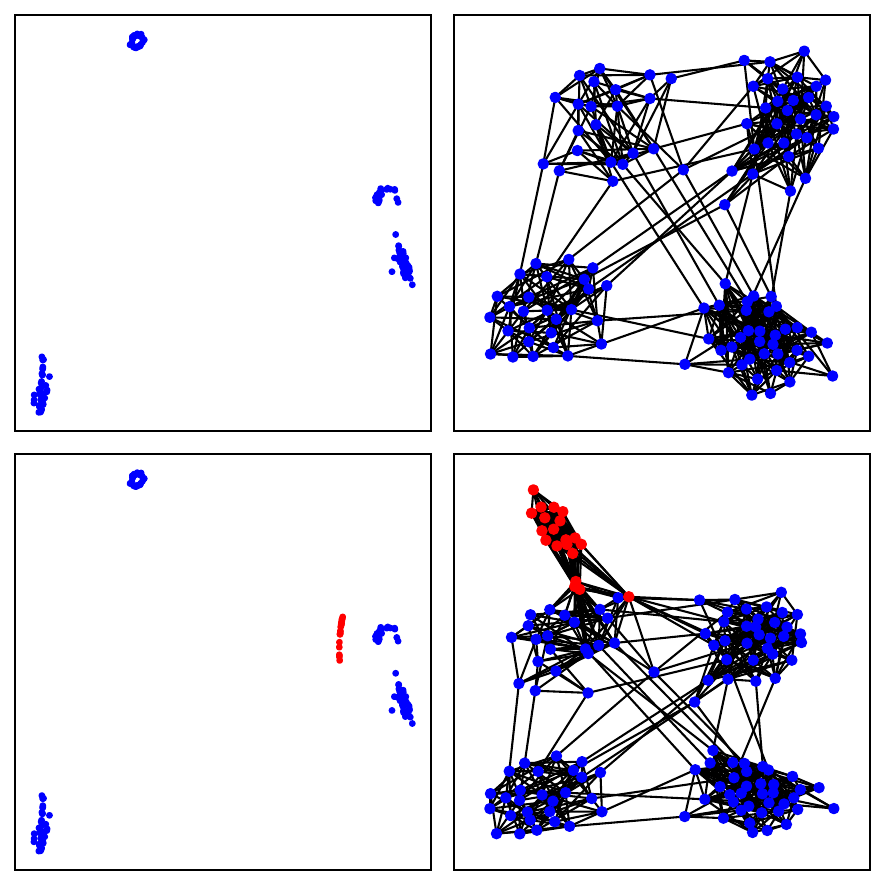}
\caption{An example of the data points in the latent space and a generated graph for SBM dataset. The blue and red points indicate the ones in the original graph and the generated one, respectively.}
\label{fig:plot_example_sbm}
\end{figure}

We further visualized what GCA was able to generate.  
Fig.~\ref{fig:plot_example_sbm} shows a sample of the generated graphs and their latent representation for %Planar dataset. 
SBM dataset. 
We projected the latent variables in a high dimensional latent space to data points in two dimensional space with $t$-SNE~\cite{Maaten2008JMLR}. 
%and UMAP~\cite{McInnes2018JOSS}. 
The original latent variables and the original graph are shown 
in the upper panel in Fig.~\ref{fig:plot_example_sbm}, 
whereas the generated latent variables and the generated graph are shown in the lower panel.  
We observed that the generated nodes were  separated from the original ones in the latent space. 
Therefore, we confirmed  that they were \textit{novel} in the sense that they were  significantly different from the original ones. 
Meanwhile, we also observed  that the generated nodes were embedded near the original graph. 
It indicates that the generated graphs do not destroy the original graph structure but rather inherits the nature of the original graph. 
In summary, 
GCA was able to generate a  novel graph community inheriting the basic nature in the original graph. %$ but having the  low probabilities of generation. 
%in a reasonable computation time. 

\subsection{Real Dataset~(RQ1)}
\label{subsection:real_dataset}

\subsubsection{Data}
\label{subsubsection:real_dataset_data}

We used the following four datasets: 
Cora~(2,708 nodes, 1,433 node features, and 7 classes), 
PubMed~(19,717 nodes, 500 node features, and 3 classes), 
CiteSeer~(3,327 nodes and 3,703 node features, 6 classes), 
%Cora-full~(19,793 nodes), 
%Coauthor-CS~(18,333 nodes), 
and Coauthor-Physics~(34,493 nodes, 495,924 edges, 5 classes). 
We used all the four real datasets with PyTorch-Geometric library\footnote{\url{https://pytorch-geometric.readthedocs.io/}}. 
%We also used CiteSeer~(3327 nodes and 3703 node features) in the Supplementary material. 
We split edges in each dataset into training dataset~(80\%) and test dataset~(20\%). 
%We trained GMM-based DDPM with the training dataset for edge prediction and confirm its performance with the test dataset. 

\subsubsection{Methods for Comparison}
\label{subsubsection:real_dataset_methods_for_comparison}

We chose 
GraphRNN\footnote{\url{https://github.com/JiaxuanYou/graph-generation}}~\cite{You2018ICML} 
and 
EDGE\footnote{\url{https://github.com/tufts-ml/graph-generation-edge}}~\cite{Chen2023ICML} 
for comparison 
because they can handle large-scale graphs. 
GraphRNN is based on the recurrent neural network, 
and 
EDGE is a diffusion model for large-scale graphs. 
%The detailed implementations 
%and experimental settings of GraphRNN and EDGE are summarized in Appendix. 

\subsubsection{Evaluation Metrics}
\label{subsubsection:real_dataset_evaluation_metrics}

For measuring the degree of extrapolation search, 
we employed Anomaly, in the same way as in the synthetic dataset in Sec.~\ref{subsection:synthetic_dataset}. 
We did not employ V.U.N. %because it is not reasonable to calculate V.U.N. 
because there is only one graph in Cora,  PubMed, CiteSeer, and Coauthor-Physics datasets. 
By contrast, for measuring the degree of maintenance of structure of
original graphs, 
we again used Deg., Clus., Orbit, and Spec., 
just in the same way as in the synthetic dataset in Sec.~\ref{subsection:synthetic_dataset}. 
Anomaly, Deg., Clus., Orbit, and Spec. were calculated in reference to the original graph in each dataset. 
%we followed the evaluation metrics in previous studies~\cite{Thiagarajan2021NeurIPS, 
%Chen2023ICML}; 
%power-law exponent of the degree sequence~(PLE), 
%normalized triangle counts~(NTC), 
%global clustering coefficient~(CC)~\cite{Chanpuriya2021NeurIPS}, 
%characteristic path
%length~(CPL), 
%and assortativity coefficient~(AC)~\cite{Newman2002PRL}. 

\subsubsection{Hyperparameter Setting}

We followed the hyperparameter setting for the synthetic dataset in Sec.~\ref{subsubsection:hyperparameter_setting_synthetic_dataset}; the candidate set of the number of clusters $\mathcal{K} = \{ 2, \dots, 10 \}$, 
the candidate set of the embedding dimension $\mathcal{D}=\{4, 8, 16, 24, 32\}$. 
We changed the number of nodes added in the new component among $M \in \{ 50, 100, 200 \}$. 
The architecture of VGAE was the same as one in the synthetic dataset in Sec.~\ref{section:synthetic_dataset_data}. 

\subsubsection{Results}
\label{subsubsection:real_dataset_results}

\begin{figure}[tb]
\centering
\includegraphics[width=0.4\linewidth]{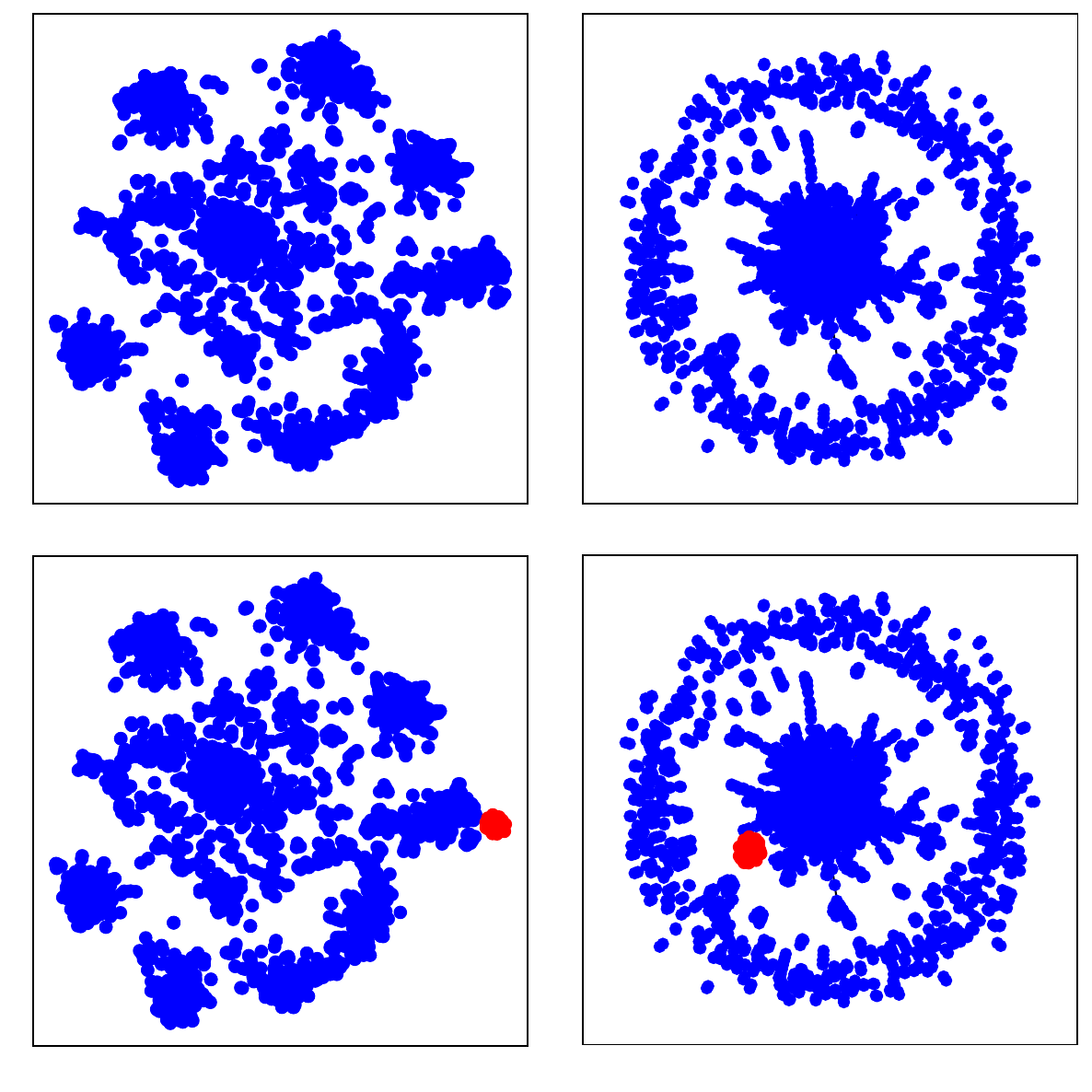}
\caption{The data points in the latent space and the generated graph for CiteSeer dataset. The blue and red points indicate the ones in the original graph and the generated one, respectively.}
\label{fig:plot_example_citeseer}
\end{figure}

Table~\ref{table:evaluation_results_on_the_real_datasets} shows the evaluation results on the real datasets. 
We observed that GCA  generated novel graphs in terms of Anomaly, compared to GraphRNN and EDGE. 
We also observed that GCA was comparable to 
%the compared algorithms for maintaining the original structure of graphs to some extent. 
the rival algorithms for maintaining the original nature of graphs to some extent. 
Fig.~\ref{fig:plot_example_citeseer} shows the generated graph and its latent representation for CiteSeer dataset. 
Again, we visualized what GCA generated by projecting the data points in the latent space into two dimensional space. 
We observed from the lower panel that there was a new cluster in the latent space with GCA, and the nodes in the new cluster were not far away from the existing nodes in the generated graph. 
In summary, 
we generated novel graphs while keeping the essential nature of the original graphs in a reasonable computation time. 

\begin{table*}[tb]
\caption{Evaluation results on the real datasets.}
\label{table:evaluation_results_on_the_real_datasets}
\begin{footnotesize}
\centering
{\tabcolsep=0.2\tabcolsep
\begin{tabular}{lrllllrrrllllrr}
\toprule
& 
\multicolumn{7}{c}{Cora} &
\multicolumn{7}{c}{PubMed} \\
\midrule
\multicolumn{1}{c}{} & 
\multicolumn{1}{c}{Anomaly} & 
\multicolumn{1}{c}{Deg.} & 
\multicolumn{1}{c}{Clus.} & 
\multicolumn{1}{c}{Orbit} & 
\multicolumn{1}{c}{Spec.} & 
\multicolumn{1}{c}{Time(T)} &
\multicolumn{1}{c}{Time(A)} &
\multicolumn{1}{c}{Anomaly} & 
\multicolumn{1}{c}{Deg.} & 
\multicolumn{1}{c}{Clus.} & 
\multicolumn{1}{c}{Orbit} & 
\multicolumn{1}{c}{Spec.} & 
\multicolumn{1}{c}{Time(T)} &
\multicolumn{1}{c}{Time(A)}
\\
\midrule
%ERGMS &
%      &
%      &
%      &
%      &
%      &
%      &  % Cora 
%      &
%      &
%      &
%      &
%      &
%      &  % PubMed 
%\\
%GPT-GNN &
%        &
%        &
%        &
%        &
%        &
%        &  % Cora 
%        &
%        &
%        &
%        &
%        &
%        &
%        &  % PubMed 
%\\
GraphRNN & 
$0.94 \pm 0.17$ & 
$0.0512$ & 
$0.0539$ & 
$0.3612$ & 
$\underline{0.0962}$ & 
115m & % Cora 
8m56s
&
$1.33 \pm 0.31$ & 
$0.0391$ & 
$0.0459$ &
$0.3235$ & 
$0.1259$ &
$163$m &  % PubMed
13m14s
\\
EDGE & 
$1.13 \pm 0.05$ & 
$\mathbf{0.0123}$ & $\mathbf{0.0352}$ & $\mathbf{0.1721}$ & $\mathbf{0.0813}$ & 
\underline{65m} &  % Cora
6m45s
&
$1.59 \pm 0.26$ & 
$\mathbf{0.0239}$ & 
$\mathbf{0.0258}$ & 
$\mathbf{0.1593}$ & 
$\mathbf{0.0759}$ & 
\underline{105m} & 
10m08s % PubMed 
\\
GCA~($M=50$) & 
$6.39 \pm 0.33$ & 
$\underline{0.0231}$ & 
$\underline{0.0437}$ & 
$\underline{0.2352}$ & 
$0.0983$ & 
$\mathbf{52m}$ & $\mathbf{28s}$ % Cora
& 
$\underline{8.69\pm 0.37}$ & 
$\underline{0.0368}$ & 
$\underline{0.0311}$ & 
$\underline{0.2359}$ & 
$\underline{0.1112}$ & 
$\mathbf{93m}$ & $\mathbf{57s}$ % PubMed
\\ 
GCA~($M=100$) & 
$\underline{6.57 \pm 0.32}$ & 
$0.0298$ & 
$0.0513$ & 
$0.2519$ & 
$0.1021$ & 
$\mathbf{52m}$ & $\mathbf{28s}$ % Cora
& 
$8.47 \pm 0.43$ & 
$0.0399$ & 
$0.0613$ &
$0.3319$ & 
$0.1459$ & 
$\mathbf{93m}$ & 
\underline{1m04s}  % PubMed
\\  
GCA~($M=200$) & 
$\mathbf{6.66 \pm 0.35}$ & 
$0.0311$ & 
$0.0535$ & 
$0.2633$ & 
$0.1391$ & 
$\mathbf{52m}$ & \underline{31s} % Cora 
& 
$\mathbf{8.98 \pm 0.28}$ & 
$0.0427$ & 
$0.0615$ & 
$0.3591$ & 
$0.1632$ & 
$\mathbf{93m}$ & 
1m08s  %PubMed
\\  
\midrule
& 
\multicolumn{7}{c}{CiteSeer} &
\multicolumn{7}{c}{Coauthor-Physics} \\
\midrule
\multicolumn{1}{c}{} & 
\multicolumn{1}{c}{Anomaly} & 
\multicolumn{1}{c}{Deg.} & 
\multicolumn{1}{c}{Clus.} & 
\multicolumn{1}{c}{Orbit} & 
\multicolumn{1}{c}{Spec.} & 
\multicolumn{1}{c}{Time(T)} &
\multicolumn{1}{c}{Time(A)} &
\multicolumn{1}{c}{Anomaly} & 
\multicolumn{1}{c}{Deg.} & 
\multicolumn{1}{c}{Clus.} & 
\multicolumn{1}{c}{Orbit} & 
\multicolumn{1}{c}{Spec.} & 
\multicolumn{1}{c}{Time(T)} &
\multicolumn{1}{c}{Time(A)}
\\
\midrule
%ERGMS &
%      &
%      &
%      &
%      &
%      &
%      &  % Cora 
%      &
%      &
%      &
%      &
%      &
%      &  % PubMed 
%\\
%GPT-GNN &
%        &
%        &
%        &
%        &
%        &
%        &  % Cora 
%        &
%        &
%        &
%        &
%        &
%        &
%        &  % PubMed 
%\\
GraphRNN & 
$1.03 \pm 0.23$ & 
$\underline{0.0371}$ & 
$\underline{0.0213}$ & 
$0.2861$ & 
$0.1033$ & 
131m & 8m43s % Cora 
& 
$1.59 \pm 0.28$ & 
$0.4302$ & 
$0.4053$ &
$0.8493$ & 
$0.8932$ &
312m & 20m29s % PubMed
\\
EDGE & 
$1.68 \pm 0.43$ & 
$\mathbf{0.0351}$ & 
$\mathbf{0.0197}$ & 
$\mathbf{0.1236}$ & 
$\mathbf{0.0756}$ & 
\underline{76m} & 5m56s  % Cora
& 
$1.59 \pm 0.26$ & 
$\mathbf{0.3909}$ & 
$\mathbf{0.3511}$ & 
$\mathbf{0.6126}$ & 
$\mathbf{0.6365}$ & 
\underline{305m} & 22m28s % PubMed 
\\
GCA~($M=50$) & 
$5.88 \pm 0.43$ & 
$0.0389$ & 
$0.0417$ & 
$\underline{0.2283}$ & 
$\underline{0.0957}$ & 
$\mathbf{63m}$ & $\mathbf{38s}$ % Cora
& 
$\underline{10.55\pm 1.69}$ & 
$\underline{0.4058}$ & 
$\underline{0.3821}$ & 
$\underline{0.7439}$ & 
$\underline{0.7058}$ & 
$\mathbf{158m}$ & 
$\mathbf{2m01s}$ % PubMed
\\ 
GCA~($M=100$) & 
$\underline{6.05 \pm 0.35}$ & 
$0.0401$ & 
$0.0572$ & 
$0.3211$ & 
$0.1332$ & 
$\mathbf{63m}$ & $\mathbf{38s}$  % Cora
& 
$\mathbf{10.39 \pm 1.88}$ & 
$0.4429$ & 
$0.4368$ &
$0.7655$ & 
$0.7928$ & 
$\mathbf{158m}$ & 
\underline{2m10s}  % PubMed
\\  
GCA~($M=200$) & 
$\mathbf{6.66 \pm 0.35}$ & 
$0.0442$ & 
$0.0631$ & 
$0.3332$ & 
$0.1159$ & 
$\mathbf{63m}$ & \underline{41s} 
& 
$13.02 \pm 2.10$ & 
$0.4560$ & 
$0.4785$ & 
$0.8027$ & 
$0.8565$ & 
$\mathbf{158m}$ & 
2m15s %PubMed
\\ 
\bottomrule
\end{tabular}
}

%\begin{subtable}{\linewidth}
%\caption{CiteSeer}
%\centering
%{\tabcolsep=0.35\tabcolsep
%\begin{tabular}{lrllllrr}
%\toprule
%\multicolumn{1}{c}{} & 
%\multicolumn{1}{c}{Anomaly} & 
%\multicolumn{1}{c}{Deg.} & 
%\multicolumn{1}{c}{Clus.} & 
%\multicolumn{1}{c}{Orbit} & 
%\multicolumn{1}{c}{Spec.} & 
%\multicolumn{1}{c}{Train. Time} \\
%\midrule
%GraphRNN & $1.03 \pm 0.23$ & $\underline{0.0371}$ & $\underline{0.0213}$ & $0.2861$ & $0.1033$ & 131m \\
%EDGE & 
%$1.11 \pm 0.31$ & $\mathbf{0.0351}$ & $\mathbf{0.0197}$ & $\mathbf{0.1236}$ & $\mathbf{0.0756}$ & 76m \\
%NGG-DM~($M=50$) & $5.88 \pm 0.43$ & $0.0389$ & $0.0417$ & $\underline{0.2283}$ & $\underline{0.0957}$ & 63m \\ 
%NGG-DM~($M=100$) & $\underline{6.05 \pm 0.35}$ & $0.0401$ & $0.0572$ & $0.3211$ & $0.1332$ & 63m \\  
%NGG-DM~($M=200$) & $\mathbf{6.66 \pm 0.35}$ & $0.0442$ & $0.0631$ & $0.3332$ & $0.1159$ & 63m \\  
%\bottomrule
%\end{tabular}
%}
%\end{subtable}
\end{footnotesize}
\end{table*}

\subsection{Sensitivity Analysis on Hyperparameters~(RQ2)}
\label{subsection:sensitivity_analysis_on_hyperperamters}

\begin{table*}[tb]
\caption{Results of the sensitivity analysis on hyperparameters.}
\label{table:results_of_the_sensitivity_analysis_on_hyperparameters}
\begin{footnotesize}
\centering
{\tabcolsep=0.7\tabcolsep
\begin{tabular}{lrllllrllll}
\toprule
& 
\multicolumn{5}{c}{SBM} &
\multicolumn{5}{c}{Cora} \\
\midrule
\multicolumn{1}{c}{} & 
\multicolumn{1}{c}{Anomaly} & 
\multicolumn{1}{c}{Deg.} & 
\multicolumn{1}{c}{Clus.} & 
\multicolumn{1}{c}{Orbit} & 
\multicolumn{1}{c}{Spec.} & 
\multicolumn{1}{c}{Anomaly} & 
\multicolumn{1}{c}{Deg.} & 
\multicolumn{1}{c}{Clus.} & 
\multicolumn{1}{c}{Orbit} & 
\multicolumn{1}{c}{Spec.} 
\\
\midrule
%$\delta_{0} = 0.5$ & 
%$2.11 \pm 0.38$ & 
%$0.0022$ & 
%$0.0377$ & 
%$0.0352$ & 
%$0.0238$ & 
%$1.43 \pm 0.25$ & 
%$0.0119$ & 
%$0.0293$ &
%$0.1159$ & 
%$0.0579$ 
%\\
$\delta_{0} = 1$ & 
$4.53 \pm 0.41$ & 
$0.0037$ & 
$0.0498$ & 
$0.0439$ & 
$0.0348$ & 
$3.64 \pm 0.55$ & 
$0.0152$ & 
$0.0343$ &
$0.1599$ & 
$0.0778$ 
\\
$\delta_{0}=5$ & 
$5.78 \pm 0.74$ & 
$0.0009$ & 
$0.0587$ & 
$0.0732$ & 
$0.0412$ & 
$6.39\pm 0.33$ & 
$0.0231$ & 
$0.0437$ & 
$0.2352$ & 
$0.0983$  
\\ 
$\delta_{0}=10$ & 
$7.55 \pm 0.36$ & 
$0.0076$ & 
$0.0832$ & 
$0.0789$ & 
$0.0592$ & 
$9.12 \pm 0.36$ & 
$0.0405$ & 
$0.0604$ &
$0.3586$ & 
$0.1593$  
\\  
%$\delta_{0}=50$ & 
%$9.37 \pm 0.35$ & 
%$0.0121$ & 
%$0.0145$ & 
%$0.1193$ & 
%$0.0923$ & 
%$11.15 \pm 0.43$ & 
%$0.0533$ & 
%$0.1193$ &
%$0.5132$ & 
%$0.3197$  
%\\ 
\midrule
$\delta_{1}=0.5D_{\max}$ & 
$2.63 \pm 0.41$ & 
$0.0023$ & 
$0.0512$ & 
$0.0412$ & 
$0.0275$ & 
$4.45 \pm 0.61$ & 
$0.0198$ & 
$0.0405$ &
$0.1995$ & 
$0.0891$ 
\\
$\delta_{1}=D_{\max}$ & 
$5.78 \pm 0.74$ & 
$0.0009$ & 
$0.0587$ & 
$0.0732$ & 
$0.0412$ & 
$6.39\pm 0.33$ & 
$0.0231$ & 
$0.0437$ & 
$0.2352$ & 
$0.0983$  
\\
$\delta_{1}=2 D_{\max}$ & 
$8.69 \pm 0.38$ & 
$0.0071$ & 
$0.0893$ & 
$0.0911$ & 
$0.0845$ & 
$10.38 \pm 0.37$ & 
$0.0412$ & 
$0.0665$ &
$0.3059$ & 
$0.1126$ 
\\
%$\delta_{1}=10 D_{\max}$ & 
%$11.19 \pm 0.24$ & 
%$0.0149$ & 
%$0.1395$ & 
%$0.1263$ & 
%$0.1261$ & 
%$12.13 \pm 0.41$ & 
%$0.0515$ & 
%$0.0853$ &
%$0.3922$ & 
%$0.1446$ 
%\\
\bottomrule
\end{tabular}
}

%\begin{subtable}{\linewidth}
%\caption{CiteSeer}
%\centering
%{\tabcolsep=0.35\tabcolsep
%\begin{tabular}{lrllllrr}
%\toprule
%\multicolumn{1}{c}{} & 
%\multicolumn{1}{c}{Anomaly} & 
%\multicolumn{1}{c}{Deg.} & 
%\multicolumn{1}{c}{Clus.} & 
%\multicolumn{1}{c}{Orbit} & 
%\multicolumn{1}{c}{Spec.} & 
%\multicolumn{1}{c}{Train. Time} \\
%\midrule
%GraphRNN & $1.03 \pm 0.23$ & $\underline{0.0371}$ & $\underline{0.0213}$ & $0.2861$ & $0.1033$ & 131m \\
%EDGE & 
%$1.11 \pm 0.31$ & $\mathbf{0.0351}$ & $\mathbf{0.0197}$ & $\mathbf{0.1236}$ & $\mathbf{0.0756}$ & 76m \\
%NGG-DM~($M=50$) & $5.88 \pm 0.43$ & $0.0389$ & $0.0417$ & $\underline{0.2283}$ & $\underline{0.0957}$ & 63m \\ 
%NGG-DM~($M=100$) & $\underline{6.05 \pm 0.35}$ & $0.0401$ & $0.0572$ & $0.3211$ & $0.1332$ & 63m \\  
%NGG-DM~($M=200$) & $\mathbf{6.66 \pm 0.35}$ & $0.0442$ & $0.0631$ & $0.3332$ & $0.1159$ & 63m \\  
%\bottomrule
%\end{tabular}
%}
%\end{subtable}
\end{footnotesize}
\end{table*}

We then confirmed the dependency of accuracy of GCA on the hyperparameters 
%$T$, 
$\delta_{0}$ and $\delta_{1}$. 
%We changed $T \in \{ 8, 10, 12, 15, 20 \}$, 
We changed 
%$\delta_{0} \in \{ 0.5, 1, 5, 10, 50, 100 \}$ 
$\delta_{0} \in \{ 1, 5, 10 \}$ 
and 
%$\delta_{1} \in \{ D_{\max}$, $2D_{\max}$, $5D_{\max}$, $10D_{\max} \}$,   
$\delta_{1} \in \{ 0.5D_{\max}$, $D_{\max}$, $2D_{\max} \}$,   
where $D_{\max}$ is the maximum distance between the existing clusters. 
For SBM and Cora datasets, 
we investigated the following sensitivity analysis: 
%(i) influence of $\delta_{0}$ given $\delta_{1} = 2D_{\max}$, 
(i) influence of $\delta_{0}$ given $\delta_{1} = D_{\max}$, 
and 
(ii) influence of $\delta_{1}$ given $\delta_{0}=5$. 
The number of generated nodes was set to $M=5$ for SBM dataset and $M=50$ for Cora dataset. 

Table~\ref{table:results_of_the_sensitivity_analysis_on_hyperparameters} shows the results of the sensitivity analysis. 
We observed that the anomaly scores increased as both $\delta_{0}$ and $\delta_{1}$ increased, 
while Deg., Clus., Orbit, and Spec. did in the same trend. 
This observation coincided with our expectation. 
The results of Table~\ref{table:results_of_the_sensitivity_analysis_on_hyperparameters} show that a wide variety of graphs with a new community structure was generated. 
In that sense, 
the results suggest that we should first specify   
 ranges of $\delta_{0}$ and  $\delta_{1}$ 
and then select novel graphs with  new communities 
by referring to the metrics such as the Anomaly score, Deg., Clus., Orbit, and Spec., 
as explained  in Sec.~\ref{subsection:community_augmentation_step}.

\subsection{Ablation Study~(RQ3)}
\label{subsection:ablation_study}

Finally, we confirmed the role and effectiveness of each part of GCA,  
in particular, 
VGAE, GMM, and community augmentation~(CA). 
We further confirmed the effects by replacing VGAE with more sophisticated 
(semi-)supervised learning algorithm called GMMDA~\cite{Li2023ICDM}; 
GMMDA jointly learns graph neural network~(GNN) for embedding a graph into latent space, 
GMM, 
and decoder. 
Note that GMMDA receives a graph $G = (X, A, y)$, 
where $X \in \mathbb{R}^{N \times F}$ is a node feature matrix, 
$A \in \mathbb{R}^{N \times N}$ is a weighted adjacency matrix, 
and 
$y \in \{ 1, \dots, K \}^{N}$ is a set of node labels. 
In addition, 
we also confirmed whether \textit{denoising} in the latent space affected the performance of GCA. 
We used the GMM-based diffusion model~\cite{Regol2023AAAI} called the Gaussian mixture model for categorical data~(GMCD),  
which employs the denoising diffusion probabilistic model~\cite{Ho2020NeurIPS}. 
%We 
%Specifically, 
%we conducted ablation study by comparing the accuracies 
%among 
%(i) GCA with GMM and diffusion model, 
%(ii) GCA with GMM alone, 
%and 
%(iii) GCA with diffusion model. 
To this end, 
we set the following experimental settings for SBM and Cora datasets:
\begin{enumerate}
\item (i) VGAE + (ii) GMM + (iii) CA~($=$GCA)
\item (i) VGAE + (ii) GMM 
\item (i) GMMDA + (ii) CA  
\item (i) VGAE + (ii) GMM + (iii) GMCD + (iv) CA
\item (i) GMMDA + (ii) GMCD + (iii) CA
\end{enumerate}
For each configuration, 
we sample $M=5$ points for SBM dataset 
and $M=50$ points for Cora dataset, 
and then compared the performance. 
We used the implementations of 
GMMDA\footnote{\url{https://anonymous.4open.science/r/GMMDA-A1C0/}} 
and 
GMCD\footnote{\url{https://github.com/networkslab/gmcd}}. 

\begin{table*}[tb]
\caption{Results of the ablation study.}
\label{table:results_of_the_ablation_study}
\begin{footnotesize}
\centering
{\tabcolsep=0.5\tabcolsep
\begin{tabular}{lrllllrrrllllrr}
\toprule
& 
\multicolumn{5}{c}{SBM} &
\multicolumn{5}{c}{Cora} \\
\midrule
\multicolumn{1}{c}{} & 
\multicolumn{1}{c}{Anomaly} & 
\multicolumn{1}{c}{Deg.} & 
\multicolumn{1}{c}{Clus.} & 
\multicolumn{1}{c}{Orbit} & 
\multicolumn{1}{c}{Spec.} & 
\multicolumn{1}{c}{Anomaly} & 
\multicolumn{1}{c}{Deg.} & 
\multicolumn{1}{c}{Clus.} & 
\multicolumn{1}{c}{Orbit} & 
\multicolumn{1}{c}{Spec.} & 
\\
\midrule
1) (i) VGAE + (ii) GMM + (iii) CA~(GCA) & 
$5.78 \pm 0.74$ & 
$0.0009$ & 
$0.0587$ & 
$0.0732$ & 
$0.0412$ & 
$6.39 \pm 0.33$ & 
$0.0231$ & 
$0.0437$ &
$0.2352$ & 
$0.0983$ &
\\
2) (i) VGAE + (ii) GMM & 
$1.39 \pm 0.13$ & 
$0.0010$ & 
$0.0331$ & 
$0.0265$ & 
$0.0071$ & 
$0.69\pm 0.37$ & 
$0.0368$ & 
$0.0311$ & 
$0.2359$ & 
$0.1112$ & 
\\ 
3) (i) GMMDA + (ii) CA & 
$6.21 \pm 0.48$ & 
$0.0059$ & 
$0.0655$ & 
$0.0523$ & 
$0.0453$ & 
$1.59 \pm 0.26$ & 
$0.0239$ & 
$0.0258$ & 
$0.1593$ & 
$0.0759$ & 
\\
4) (i) VGAE + (ii) GMM + (iii) GMCD + (iv) CA & 
$5.69 \pm 0.51$ & 
$0.0059$ & 
$0.0638$ & 
$0.0547$ & 
$0.0428$ & 
$0.66 \pm 0.33$ & 
$0.0382$ & 
$0.0295$ &
$0.2515$ & 
$0.1339$ & 
\\
5) (i) GMMDA + (ii) GMCD + (iii) CA & 
$6.17 \pm 0.38$ & 
$0.0065$ & 
$0.0647$ & 
$0.0545$ & 
$0.0439$ & 
$1.55 \pm 0.41$ & 
$0.0261$ & 
$0.0244$ &
$0.1822$ & 
$0.1197$ & 
\\  
\bottomrule
\end{tabular}
}

%\begin{subtable}{\linewidth}
%\caption{CiteSeer}
%\centering
%{\tabcolsep=0.35\tabcolsep
%\begin{tabular}{lrllllrr}
%\toprule
%\multicolumn{1}{c}{} & 
%\multicolumn{1}{c}{Anomaly} & 
%\multicolumn{1}{c}{Deg.} & 
%\multicolumn{1}{c}{Clus.} & 
%\multicolumn{1}{c}{Orbit} & 
%\multicolumn{1}{c}{Spec.} & 
%\multicolumn{1}{c}{Train. Time} \\
%\midrule
%GraphRNN & $1.03 \pm 0.23$ & $\underline{0.0371}$ & $\underline{0.0213}$ & $0.2861$ & $0.1033$ & 131m \\
%EDGE & 
%$1.11 \pm 0.31$ & $\mathbf{0.0351}$ & $\mathbf{0.0197}$ & $\mathbf{0.1236}$ & $\mathbf{0.0756}$ & 76m \\
%NGG-DM~($M=50$) & $5.88 \pm 0.43$ & $0.0389$ & $0.0417$ & $\underline{0.2283}$ & $\underline{0.0957}$ & 63m \\ 
%NGG-DM~($M=100$) & $\underline{6.05 \pm 0.35}$ & $0.0401$ & $0.0572$ & $0.3211$ & $0.1332$ & 63m \\  
%NGG-DM~($M=200$) & $\mathbf{6.66 \pm 0.35}$ & $0.0442$ & $0.0631$ & $0.3332$ & $0.1159$ & 63m \\  
%\bottomrule
%\end{tabular}
%}
%\end{subtable}
\end{footnotesize}
\end{table*}

Table~\ref{table:results_of_the_ablation_study} shows the results of the ablation study. 
By comparing the anomaly scores in 1) and 2), 
we observed that 
CA was essential in generating graphs with a new community structure. 
Note that Deg., Clus., Orbit, and Spec. of 2) in SBM dataset were approximately equal to the ones between training and test datasets in Table~\ref{table:evaluation_results_on_the_synthetic_datasets}. 
This indicates that graphs with the same community structure as training dataset were generated without CA. 
By comparing 1) and 3), 
we confirmed that joint learning including (semi-) supervised learning with class labels of nodes was more effective than simply embedding with VGAE, 
for locating each cluster. 
By comparing 1) and 4), and 3) and 5), 
we observed that the anomaly scores slightly decreased with GMCD. However, there was relatively little difference 
and it might be in a range of randomness in the community augmentation step.  

\section{Conclusion}
\label{section:conclusion}

%This study addressed the issue of graph novelty generation with diffusion models. 
This study has addressed the issue of graph community augmentation. 
To this end, we proposed a novel mathematical framework for community augmentation to develop an effective and efficient algorithm called GCA. 
The key idea of GCA is to fit GMM to data points in the latent space mapped from a graph, and then to add a cluster there for generating  a new community under novelty and reliability conditions. 
It is the first work to formalize graph community-level knowledge extrapolation.
%We leveraged the expressiveness of the diffusion model and generated novel graphs. 
Experimental results showed the effectiveness of GCA. 
%There are several directions for extending our framework and algorithm. 
Future work includes the determination of initial values and extension of the update and stopping rules in executing novelty and reliability conditions.  

%\newpage 

\bibliographystyle{IEEEtran}
\bibliography{main}

\begin{thebibliography}{10}
\providecommand{\url}[1]{#1}
\csname url@samestyle\endcsname
\providecommand{\newblock}{\relax}
\providecommand{\bibinfo}[2]{#2}
\providecommand{\BIBentrySTDinterwordspacing}{\spaceskip=0pt\relax}
\providecommand{\BIBentryALTinterwordstretchfactor}{4}
\providecommand{\BIBentryALTinterwordspacing}{\spaceskip=\fontdimen2\font plus
\BIBentryALTinterwordstretchfactor\fontdimen3\font minus
  \fontdimen4\font\relax}
\providecommand{\BIBforeignlanguage}[2]{{%
\expandafter\ifx\csname l@#1\endcsname\relax
\typeout{** WARNING: IEEEtran.bst: No hyphenation pattern has been}%
\typeout{** loaded for the language `#1'. Using the pattern for}%
\typeout{** the default language instead.}%
\else
\language=\csname l@#1\endcsname
\fi
#2}}
\providecommand{\BIBdecl}{\relax}
\BIBdecl

\bibitem{Bond-Taylor2022TPAMI}
S.~Bond-Taylor, A.~Leach, Y.~Long, and C.~G. Willcocks, ``Deep generative
  modelling: a comparative review of {VAE}s, {GAN}s, normalizing flows,
  energy-based and autoregressive models,'' \emph{IEEE Transactions on Pattern
  Analysis and Machine Intelligence}, vol.~44, pp. 7327--7347, 2022.

\bibitem{Bonifati2020CSUR}
A.~Bonifati, I.~Holubová, A.~Prat-Pérez, and S.~Sakr, ``Graph generators:
  state of the art and open challenges,'' \emph{ACM Computing Surveys},
  vol.~53, no.~2, pp. 1--30, 2020.

\bibitem{Faez2021IEEEAccess}
F.~Faez, Y.~Ommi, M.~S. Baghshah, and H.~R. Rabiee, ``Deep graph generators: a
  survey,'' \emph{IEEE Access}, vol.~9, pp. 106\,675--106\,702, 2021.

\bibitem{Guo2022TPAMI}
X.~Guo and L.~Zhao, ``A systematic survey on deep generative models for graph
  generation,'' \emph{IEEE Transactions on Pattern Analysis and Machine
  Intelligence}, vol.~45, no.~5, pp. 5370--5390, 2022.

\bibitem{Zhu2022LoG}
Y.~Zhu, Y.~Du, Y.~Wang, Y.~Xu, J.~Zhang, Q.~Liu, and S.~Wu, ``A survey on deep
  graph generation: methods and applications,'' in \emph{Proc. of LoG}, 2022.

\bibitem{Ding2022SIGKDDExplorations}
K.~Ding, Z.~Xu, H.~Tong, and H.~Liu, ``Data augmentation for deep graph
  learning: a survey,'' \emph{ACM SIGKDD Explorations Newsletter}, vol.~24,
  no.~2, pp. 61--77, 2022.

\bibitem{Zhao2023IDEB}
T.~Zhao, W.~Jin, Y.~Liu, Y.~Wang, G.~Liu, S.~Günneman, N.~Shah, and M.~Jiang,
  ``Graph data augmentation for graph machine learning: a survey,'' \emph{IEEE
  Data Engineering Bulletin}, vol.~46, no.~2, pp. 140--165, 2023.

\bibitem{Ohsawa1998ADL}
Y.~Ohsawa, N.~E. Benson, and M.~Yachida, ``Keygraph: automatic indexing by
  co-occurrence graph based on building construction metaphor,'' in
  \emph{Proceedings IEEE International Forum on Research and Technology
  Advances in Digital Libraries~(ADL)}, 1998, pp. 12--18.

\bibitem{Ohsawa2002NGC}
Y.~Ohsawa, ``Chance discoveries for making decisions in complex real world,''
  \emph{New Generation Computing}, no.~20, pp. 143--163, 2002.

\bibitem{Harris2014book}
J.~K. Harris, \emph{An introduction to exponential random graph
  modeling}.\hskip 1em plus 0.5em minus 0.4em\relax SAGE Publications, 2014.

\bibitem{Cavallari2017CIKM}
S.~Cavallari, V.~W. Zheng, H.~Cai, K.~C.-C. Chang, and E.~Cambria, ``Learning
  community embedding with community detection and node embedding on graphs,''
  in \emph{Proc. of CIKM}, 2017, pp. 377--386.

\bibitem{Yang2019ICCV}
L.~Yang, N.-M. Cheung, J.~Li, and J.~Fang, ``Deep clustering by {G}aussian
  mixture variational autoencoders with graph embedding,'' in \emph{Proc. of
  ICCV}, 2019, pp. 6440--6449.

\bibitem{Feng2021ICML}
R.~Feng, J.~Xiao, K.~Zheng, D.~Zhao, J.~Zhou, Q.~Sun, and Z.-J. Zha,
  ``Principled knowledge extrapolation with {GAN}s,'' in \emph{Proc. of ICML},
  2021, pp. 6447--6464.

\bibitem{Chan2021NeurIPS}
A.~Chan, A.~Madani, B.~Krause, and N.~Naik, ``Deep extrapolation for
  attribute-enhanced generation,'' in \emph{Proc. of NeurIPS}, 2021, pp.
  14\,084--14\,096.

\bibitem{Besserve2021AAAI}
M.~Besserve, R.~Sun, D.~Janzing, and B.~Scholkopf, ``A theory of independent
  mechanisms for extrapolation in generative models,'' in \emph{Proc. of AAAI},
  2021, pp. 6741--6749.

\bibitem{Li2021WACV}
Y.~Li, L.~Jiang, and M.-H. Yang, ``Controllable and progressive image
  extrapolation,'' in \emph{Proc. of WACV}, 2021, pp. 2140--2149.

\bibitem{Han2019KDD}
F.~X. Han, D.~Niu, H.~Chen, K.~Lai, Y.~He, and Y.~Xu, ``A deep generative
  approach to search extrapolation and recommendation,'' in \emph{Proc. of
  KDD}, 2019, pp. 1771--1779.

\bibitem{Azabou2023ICML}
M.~Azabou, V.~Ganesh, S.~Thakoor, C.-H. Lin, L.~Sathidevi, R.~Liu, M.~Valko,
  P.~Veličković, and E.~L. Dyer, ``Half-{H}op: a graph upsampling approach
  for slowing down message passing,'' in \emph{Proc. of ICML}, 2023, pp.
  1341--1360.

\bibitem{Liu2022ICML}
S.~Liu, R.~Ying, H.~Dong, L.~Li, T.~Xu, Y.~Rong, P.~Zhao, J.~Huang, and D.~Wu,
  ``Local augmentation for graph neural networks,'' in \emph{Proc. of ICML},
  2022, pp. 14\,054--14\,072.

\bibitem{Li2023ICDM}
Y.~Li, L.~Xu, and K.~Yamanishi, ``{GMMDA}: Gaussian mixture modeling of graph
  in latent space for graph data augmentation,'' in \emph{Proc. of ICDM}, 2023,
  pp. 319--328.

\bibitem{Lin2023ICLR}
L.~Lin, J.~Chen, and H.~Wang, ``Spectral augmentation for self-supervised
  learning on graphs,'' in \emph{Proc. of ICLR}, 2022.

\bibitem{Kim2022NeurIPS}
D.~Kim, J.~Baek, and S.~J. Hwang, ``Graph self-supervised learning with
  accurate discrepancy learning,'' in \emph{Proc. of NeurIPS}, 2022.

\bibitem{Li2022arXiv}
H.~Li, X.~Wang, Z.~Zhang, and W.~Zhu, ``Out-of-distribution generalization on
  graphs: a survey,'' arXiv preprint 2202.07987, 2022.

\bibitem{Kocaoglu2018ICLR}
M.~Kocaoglu, C.~Snyder, A.~G. Dimakis, and S.~Vishwanath, ``Causal{GAN}:
  Learning causal implicit generative models with adversarial training,'' in
  \emph{Proc. of ICLR}, 2018.

\bibitem{Sauer2020ICLR}
A.~Sauer and A.~Geiger, ``Counterfactual generative networks,'' in \emph{Proc.
  of ICLR}, 2020.

\bibitem{Nemirovsky2020arXiv}
D.~Nemirovsky, N.~Thiebaut, Y.~Xu, and A.~Gupta, ``Counte{RGAN}: Generating
  realistic counterfactuals with residual generative adversarial nets,''
  \emph{arXiv preprint arXiv:2009.05199}, 2020.

\bibitem{Yang2021CVPR}
M.~Yang, F.~Liu, Z.~Chen, X.~Shen, J.~Hao, and J.~Wang, ``Causal{VAE}:
  disentangled representation learning via neural structural causal models,''
  in \emph{Proc. of CVPR}, 2021, pp. 9593--9602.

\bibitem{Averitt2020JBiomedInform}
A.~J. Averitt, N.~Vanitchanant, R.~Ranganath, and A.~J. Perotte, ``The
  counterfactual $\chi$-{GAN}: finding comparable cohorts in observational
  health data,'' \emph{Journal of Biomedical Informatics}, vol. 109, p. 103515,
  2020.

\bibitem{Thiagarajan2021NeurIPS}
J.~J. Thiagarajan, V.~Narayanaswamy, D.~Rajan, J.~Liang, A.~Chaudhari, and
  A.~Spanias, ``Designing counterfactual generators using deep model
  inversion,'' in \emph{Proc. of NeurIPS}, vol.~34, 2021.

\bibitem{Pearl2009StatSurveys}
J.~Pearl, ``Causal inference in statistics: An overview,'' \emph{Statistics
  surveys}, vol.~3, pp. 96--146, 2009.

\bibitem{Bishop2006PRML}
C.~M. Bishop, \emph{Pattern Recognition and Machine Learning}.\hskip 1em plus
  0.5em minus 0.4em\relax Springer, 2006.

\bibitem{Zhang2022TKDE}
Z.~Zhang, P.~Cui, and W.~Zhu, ``Deep learning on graphs: a survey,'' \emph{IEEE
  Transactions on Knowledge and Data Engineering}, vol.~34, pp. 249--270, 2022.

\bibitem{Kipf2016arXiv}
T.~N. Kipf and M.~Welling, ``Variational graph auto-encoders,'' arXiv preprint
  1611.07308, 2016.

\bibitem{Rissanen1978Automatica}
J.~Rissanen, ``Modeling by shortest data description,'' \emph{Automatica},
  vol.~14, pp. 465--471, 1978.

\bibitem{Yamanishi2023Springer}
K.~Yamanishi, \emph{Learning with the Minimum Description Length
  Principle}.\hskip 1em plus 0.5em minus 0.4em\relax Springer Nature, 2023.

\bibitem{Wu2017KDD}
T.~Wu, S.~Sugawara, and K.~Yamanishi, ``Decomposed normalized maximum
  likelihood codelength criterion for selecting hierarchical latent variable
  models,'' in \emph{Proc. of KDD}, 2017, pp. 1165--1174.

\bibitem{Yamanishi2019DAMI}
K.~Yamanishi, T.~Wu, S.~Sugawara, and M.~Okada, ``The decomposed normalized
  maximum likelihood code-length criterion for selecting hierarchical latent
  variable models,'' \emph{Data Mining and Knowledge Discovery}, vol.~33, pp.
  1017--1058, 2019.

\bibitem{Fukushima2021ComplexNetworks}
S.~Fukushima, R.~Kanai, and K.~Yamanishi, ``Graph summarization with latent
  variable probabilistic models,'' in \emph{Proc. of ComplexNetworks}, 2021,
  pp. 428--440.

\bibitem{Fukushima2020ICDM}
S.~Fukushima and K.~Yamanishi, ``Detecting hierarchical changes in latent
  variable model,'' in \emph{Proc. of ICDM}, 2020, pp. 1128--1133.

\bibitem{Fukushima2023ICDM}
------, ``Balancing summarization and change detection in graph streams,'' in
  \emph{Proc. of ICDM}, 2023, pp. 1025--1030.

\bibitem{Rissanen2012Cambridge}
J.~Rissanen, \emph{Optimal estimation of parameters}.\hskip 1em plus 0.5em
  minus 0.4em\relax Cambridge, 2012.

\bibitem{Kontkanen2007}
P.~Kontkanen and P.~Myllymäki, ``A linear-time algorithm for computing the
  multinomial stochastic complexity,'' \emph{Information Processing Letters},
  vol. 103, no.~6, pp. 227--233, 2007.

\bibitem{Hershey2007ICASSP}
J.~R. Hershey and P.~A. Olsen, ``Approximating the {K}ullback {L}eibler
  divergence between {G}aussian mixture models,'' in \emph{Proc. of ICASSP},
  2007, pp. 317--320.

\bibitem{Durrieu2012ICASSP}
J.-L. Durrieu, J.-P. Thiran, and F.~Kelly, ``Lower and upper bounds for
  approximation of the {K}ullback-{L}eibler divergence between {G}aussian
  mixture models,'' in \emph{Proc. of ICASSP}, 2012, pp. 4833--4836.

\bibitem{Tsuda2005JMLR}
K.~Tsuda, G.~Ratsch, and M.~K. Warmuth, ``Matrix exponentiated gradient updates
  for on-line learning and {B}regman projection,'' \emph{Journal of Machine
  Learning Research}, vol.~6, pp. 995--1018, 2005.

\bibitem{Kivinen1997IC}
J.~Kivinen and M.~K. Warmuth, ``Exponentiated gradient versus gradient descent
  for linear predictors,'' \emph{Information and Computation}, vol. 132, no.~1,
  pp. 1--63, 1997.

\bibitem{Martinkus2022ICML}
K.~Martinkus, A.~Loukas, N.~Perraudin, and R.~Wattenhofer, ``{SPECTRE}:
  spectral conditioning helps to overcome the expressivity limits of one-shot
  graph generators,'' in \emph{Proc. of ICML}, 2022, pp. 15\,159--15\,179.

\bibitem{Jo2023arXiv}
J.~Jo, D.~Kim, and S.~J. Hwang, ``Graph generation with diffusion mixture,'' in
  \emph{Proc. of ICML}, 2024, pp. 22\,371--22\,405.

\bibitem{Snijders1997JoC}
T.~A.~B. Snijders, ``Estimation and prediction for stochastic blockmodels for
  graphs with latent block structure,'' \emph{Journal of Classification},
  vol.~14, pp. 75--100, 1997.

\bibitem{Jo2022ICML}
J.~Jo, S.~Lee, and S.~J. Hwang, ``Score-based generative modeling of graphs via
  the system of stochastic differential equations,'' in \emph{Proc. of ICML},
  2022, pp. 10\,362--10\,383.

\bibitem{Vignac2023ICLR}
C.~Vignac, I.~Krawczuk, A.~Siraudin, B.~Wang, V.~Cevher, and P.~Frossard,
  ``Digress: discrete denoising diffusion for graph generation,'' in
  \emph{Proc. of ICLR}, 2023.

\bibitem{Ho2020NeurIPS}
J.~Ho, A.~Jain, and P.~Abbeel, ``Denoising diffusion probabilistic models,'' in
  \emph{Proc. of NeurIPS}, 2020, pp. 6840--6851.

\bibitem{Maaten2008JMLR}
L.~van~der Maaten and G.~Hinton, ``Visualizing data using t-sne,''
  \emph{Journal of Machine Learning Research}, vol.~9, pp. 2579--2605, 2008.

\bibitem{You2018ICML}
J.~You, R.~Ying, X.~Ren, W.~L. Hamilton, and J.~Leskovec, ``Graphrnn:
  generating realistic graphs with deep auto-regressive models,'' in
  \emph{Proc. of ICML}, 2018, pp. 5708--5717.

\bibitem{Chen2023ICML}
H.~Chen, J.~He, X.~Han, and L.-P. Liu, ``Efficient and degree-guided graph
  generation via discrete diffusion modeling,'' in \emph{Proc. of ICML}, 2023,
  pp. 4585--4610.

\bibitem{Regol2023AAAI}
F.~Regol and M.~Coates, ``Diffusing gaussian mixtures for generating
  categorical data,'' in \emph{Proc. of AAAI}, 2023, pp. 9570--9578.

\end{thebibliography}

\appendix

\newcommand\longvar[1]{\mathchardef\UrlBreakPenalty=100
\mathchardef\UrlBigBreakPenalty=100\url{#1}}

\section{Detailed Derivation}
\label{section:detailed_derivation}

\subsection{Upper Bound of KL Divergence between GMMs}
\label{subsection:upper_bound_of_KL_divergence_between_GMMs}
KL divergence between two GMMs in Eq.~\eqref{eq:upperbound_of_distance_between_original_and_new_probabilistic_distributions} is upper-bounded, 
in reference to \cite{Hershey2007ICASSP,Durrieu2012ICASSP}, 
as follows: 
\begin{align}
d_{\mathrm{KL}}(p_{\mathrm{orig}}, p_{\mathrm{new}}) 
&\leq 
d_{\phi, \psi}(p_{\mathrm{orig}}, p_{\mathrm{new}}),
\end{align}
where 
\begin{align}
&d_{\phi, \psi}(p_{\mathrm{orig}}, 
 p_{\mathrm{new}}) \\
&\mydef 
%\sum_{k=1}^{K} \sum_{k'=1}^{K+1}
%\phi_{k' | k} 
%\log{ 
%    \frac{
%        \phi_{k' | k}
%    }{
%        \psi_{k | k'}
%    }
%} \\
-\sum_{k=1}^{K} \sum_{k'=1}^{K+1} 
    \phi_{k' | k}
    \int 
        \mathcal{N}(
            v; 
            \mu_{k}, 
            \Sigma_{k}) 
    \log{ 
        \frac{
            \psi_{k|k'} \mathcal{N}(
                v; 
                \mu_{k'}, 
                \Sigma_{k'}
            )
        }{
            \phi_{k'|k}
            \mathcal{N}(
                v;
                \mu_{k}, 
                \Sigma_{k}
            )
        }
    }
    \, \mathrm{d}v, \\
&\psi_{k | k'}
=  \frac{
        w_{k'}' \phi_{k'|k}
    }{
        \sum_{\ell} \phi_{k'|\ell}
    }, \\
&\phi_{k' | k}
= \frac{
        w_{k} \psi_{k | k'}
        \exp{(-d_{\mathrm{KL}}
        (\mathcal{N}
        (v; 
         \mu_{k}, 
         \Sigma_{k}), 
         \mathcal{N}(v; \mu_{k'}, 
         \Sigma_{k'}))}
    }{
        \sum_{\ell'}
            \psi_{k|\ell'}
            \exp{(-
            d_{\mathrm{KL}}
            (\mathcal{N}(
             v; 
             \mu_{k}, 
             \Sigma_{k}),  
             \mathcal{N}(
             v;
             \mu_{\ell'}, 
             \Sigma_{\ell'})
            )}
    }. 
%\label{eq:upperbound_of_distance_between_original_and_new_probabilistic_distributions}
\end{align}
Here, 
$\phi$ and $\psi$ satisfy 
the following equations: 
\begin{align}
p_{\mathrm{orig}} 
&= \sum_{k=1}^{K} \sum_{\ell=1}^{K+1} 
\phi_{\ell | k} \mathcal{N}(v; \mu_{k}, \Sigma_{k}), \\
p_{\mathrm{new}} 
&= \sum_{m=1}^{K} \sum_{k'=1}^{K+1} 
   \psi_{m | k'} \mathcal{N}(v; \mu_{k'}, \Sigma_{k'}), 
\end{align} 
where 
$p_{\mathrm{orig}}$ and 
$p_{\mathrm{new}}$ are 
defined in Eq.~\eqref{eq:original_probability_density} 
and 
\eqref{eq:probability_density_with_newly_generated_clusters}, respectively. 
Therefore, 
the following relations hold between $\phi$, $\psi$, $w$, and $w'$: 
%$\sum_{\ell=1}^{K+1} \phi_{\ell | k} = w_{k}$ 
%and 
%$\sum_{m=1}^{K} \psi_{m | k'} = w_{k'}'$. 
\begin{align}
\sum_{\ell=1}^{K+1} \phi_{\ell | k} = w_{k}, 
\quad 
\sum_{m=1}^{K} \psi_{m | k'} = w_{k'}'. 
\end{align}

\end{document}